\documentclass{article}
\usepackage[preprint]{neurips_2026} 

\usepackage{amsmath,amssymb,amsfonts,amsthm,mathtools}

\usepackage[utf8]{inputenc}
\usepackage[T1]{fontenc}
\usepackage{hyperref}
\usepackage{url}
\usepackage{booktabs}
\usepackage{tabularx}
\usepackage{nicefrac}
\usepackage{microtype}
\usepackage{xcolor}
\usepackage{graphicx}
\usepackage{subcaption}
\usepackage{multirow}
\usepackage{algorithm}
\usepackage{algorithmic}

\usepackage[capitalize,noabbrev]{cleveref}
\usepackage{listings}
\lstdefinelanguage{Verilog}{
  morekeywords={module,endmodule,input,output,wire,reg,assign,always,begin,end,
                if,else,case,casez,endcase,default,for,initial,parameter,localparam,
                posedge,negedge,or,and,not,xor},
  morecomment=[l]{//},
  morecomment=[s]{/*}{*/},
  morestring=[b]"
}

\theoremstyle{plain}

\theoremstyle{definition}

\theoremstyle{remark}

\crefname{assumption}{Assumption}{Assumptions}
\Crefname{assumption}{Assumption}{Assumptions}

\newcommand{\KL}{\mathrm{KL}}

\newcommand{\vtheta}{\boldsymbol{\theta}}

\newcommand{\cS}{\mathcal{S}}

\newcommand{\piTheta}{\pi_{\vtheta}}

\newcommand{\adp}{\mathrm{ADP}}

\newcommand{\ppap}{\mathrm{PPA\text{-}product}}

\newcommand{\puct}{\mathrm{PUCT}}
\newcommand{\adv}{A}
\newcommand{\Qval}{Q}
\newcommand{\Nvis}{N}
\DeclareMathOperator{\rank}{rank}
\newcommand{\betastar}{\beta^*}

\title{Alpha-RTL: Test-Time Training for RTL Hardware Optimization}

\author{%
  \bfseries Peilong Zhou$^{1,2,3}$,
  Zhirong Chen$^{1,2}$,
  Cangyuan Li$^{1,2}$,
  Haoyu Gao$^{1,2,3}$, \\[2pt]
  \bfseries Kaiyan Chang$^{1,2}$,
  Ziming Qu$^{1,2}$,
  Ying Wang$^{1,2}$ \\[6pt]
  \normalfont\small
  $^1$SKLP, Institute of Computing Technology, Chinese Academy of Sciences \\[2pt]
  $^2$University of the Chinese Academy of Sciences \quad
  $^3$School of Advanced Interdisciplinary Sciences \\[4pt]
  \textbf{Correspondence:}
  \href{mailto:wangying2009@ict.ac.cn}{\texttt{wangying2009@ict.ac.cn}}
}

\begin{document}

\maketitle

\begin{abstract}
Large language models (LLMs) have shown increasing promise in generating
functionally correct register-transfer-level (RTL) hardware designs.
Recent LLM-for-RTL systems further improve this ability through
Verilog-domain distillation, RLVR with automated testbenches, or
EDA-integrated reinforcement learning with syntax, simulation, and PPA
rewards.  However, these methods primarily train a general RTL generator
before deployment, while test-time approaches typically search with a
frozen policy.  We instead perform reinforcement learning at test time,
allowing the LLM policy to adapt to executable EDA feedback from the
specific RTL optimization problem.  This setting is challenging because
candidate designs must pass sparse discrete validity gates---syntax
checking and simulation---before receiving meaningful synthesis-derived
feedback on area, timing, or power.
We propose \textbf{TTT-RTL}, to our knowledge the first 
\emph{per-design} test-time training framework that closes the loop
between an LLM policy and an EDA pipeline for RTL optimization.
TTT-RTL samples candidate implementations, verifies them through
syntax checking and simulation, scores valid designs using
synthesis-derived PPA product, reuses high-reward variants through a
PUCT-indexed design-state pool, and updates the policy with an
entropic policy-gradient objective.  To stabilize policy updates under
sparse or plateaued reward groups, we introduce an adaptive KL-budget
controller that adjusts the entropy constraint using reference KL,
effective sample size, constant-reward fraction, and beta-search
saturation.
On RTLLM~v2.0 under Nangate 45\,nm, TTT-RTL reduces the geometric-mean
PPA product by $65.1\%$ over the reference, outperforming the strongest
published frozen-policy agent baseline under the same
reference-normalized metric at $26.1\%$.  On an industrial XuanTie
C910 FPU leading-zero-anticipation unit under Sky130, TTT-RTL achieves
a $59.4\%$ ADP reduction over the original implementation, and
ablations show that policy adaptation, state reuse, and KL-budget
control each contribute to the gain.
These results suggest that test-time training with executable EDA
feedback can move LLM-based RTL generation beyond functional
correctness toward physically optimized hardware.

\end{abstract}

\section{Introduction}

Hardware design is a laborious process whose quality is ultimately judged by
physical metrics: area, clock frequency, and power consumption.
Modern EDA flows translate a register-transfer-level (RTL) description written in
Verilog or VHDL into a physical circuit; the resulting
\emph{PPA product} (Area$\times$Delay$\times$Power, the joint figure
of merit used by RTLLM~v2.0 reports) depends not only on the logical
correctness of the code but on fine-grained micro-architectural choices---
pipeline depth, operator scheduling, signal encoding---that are difficult to predict
without running synthesis.

Recent work has shown that LLMs can generate functionally correct RTL code
\citep{liu2023verilogeval, liu2023rtlcoder, blocklove2023chipchat}.
These models learn to imitate high-quality Verilog from training corpora, and
when prompted with a hardware specification they can produce designs that pass
functional simulation.
However, functional correctness is a necessary but insufficient condition for
hardware quality.
A multiplier that passes all testbench cases may still be 30\% larger or
30\% slower than an optimized reference implementation.
Because physical synthesis metrics are computed by EDA tools at evaluation
time---not during LLM training---existing models cannot learn to minimize the
PPA product.

Existing approaches to optimize RTL fall into two families,
each capturing only half of what is needed.
\textbf{Frozen-LLM search agents}---EvolVE~\citep{evolve2025},
VeriAgent~\citep{veriagent2025}, and the current SOTA
REvolution~\citep{revolution2025}---drive an LLM through evolutionary
or reflective loops with synthesis feedback, but the weights never
update: search quality is bounded by the unchanged base policy, and
the design knowledge accumulated from EDA tools is discarded the
moment the run ends.
\textbf{Training-time RL methods}---ChipSeek~\citep{chen2025chipseek}
and EARL~\citep{earl2025}---do learn from synthesis signals, but they
ship a single amortized policy evaluated under Pass@$k$: each candidate
is an independent single-shot generation, with no mechanism to feed
EDA results from one attempt back into the next on the same design.

We argue the missing regime is to \emph{fuse search into training at
test time}: given one design and its reference, run a multi-round
search whose every rollout's EDA feedback updates the model
\emph{online for that problem}, so the policy itself deepens as the
search progresses.  Concretely, we treat Verilog candidates as nodes
in a PUCT tree whose every expansion is scored by a three-stage EDA
pipeline (syntax~$\to$~simulation~$\to$~synthesis), and we feed that
dense physical reward back into online policy-gradient updates on the
same design.  The result is a closed loop in which exploration and
training co-evolve under EDA-grounded feedback, rather than search
being grafted on top of a frozen policy.  On RTLLM~v2.0 this regime
substantially outperforms the strongest published agent baseline on
PPA-product reduction, and on a XuanTie~C910 industrial floating-point
unit it improves over well-tuned production RTL---we quantify both
below.

\paragraph{Contributions.}
\begin{enumerate}
  \item We formulate per-design RTL optimization as a test-time
        training problem, fusing PUCT-guided search into online
        policy updates driven by EDA feedback.

  \item We propose \textbf{TTT-RTL}, instantiating this regime with a
        Verilog state pool, EDA-feedback prompting, an entropic
        advantage estimator, and a three-stage syntax/simulation/
        synthesis reward, built on \texttt{verl}
        \citep{sheng2024hybridflow}.

  \item On \textbf{RTLLM~v2.0} TTT-RTL covers $48/49$ designs and
        cuts the PPA product by a geometric-mean of $65.1\%$ vs.\
        $26.1\%$ for the strongest published agent baseline
        (REvolution); on a \textbf{XuanTie~C910} industrial LZA unit
        it improves over well-tuned production RTL by $59.4\%$ ADP
        at seed~$42$, with single-seed component ablations isolating
        the contribution of the PUCT pool and the entropic estimator.
\end{enumerate}

\section{Related Work}
\label{sec:related}

\subsection{Test-time training and RL with verifiable rewards}

Test-time training (TTT) \citep{sun2020ttt} updates model parameters at
inference time using an auxiliary self-supervised objective constructed
from the test input.  \citet{yuksekgonul2026discover} extended TTT to
discrete reasoning via \textbf{PUCT-guided exploration}: a verifier
scores candidate solutions and policy-gradient updates improve the
generation policy for the specific problem at hand, enabling LLMs to
solve mathematical problems beyond their static training distribution.
TTRL \citep{zuo2025ttrl} similarly performs test-time RL using
majority-vote pseudo-labels on unlabeled data, while
\citet{snell2024scaling} and \citet{wu2024inference} scale
\textbf{test-time compute} without updating the model at all.
On the optimizer side, GRPO \citep{shao2024deepseekmath} provides the
group-relative policy-gradient recipe we build on, originally developed
for verifiable mathematical correctness, and CodeRL
\citep{le2022coderl} is the closest precedent in program synthesis,
using unit-test pass/fail as the RL reward signal.

\subsection{LLMs for RTL and PPA optimization}

\paragraph{Verilog generation targeting functional correctness.}
A growing body of work applies LLMs to hardware description language
generation and evaluates them on open benchmarks.  VerilogEval
\citep{liu2023verilogeval} measures pass rates on 156 HDLBits tasks,
while RTLLM \citep{lu2024rtllm} and its v2.0 successor OpenLLM-RTL
\citep{liu2025openllmrtl} extend the setting to larger designs with
synthesis feedback; RTLLM~v2.0 is the benchmark we use throughout this
paper.  On the generation side, RTLCoder \citep{liu2023rtlcoder},
Chip-Chat \citep{blocklove2023chipchat}, ChipNeMo
\citep{liu2023chipnemo}, and BetterV \citep{pei2024betterv} fine-tune
or steer LLMs for Verilog but target \textbf{functional correctness
only} and do not optimize physical metrics.

\paragraph{Frozen-policy agents for PPA.}
A more recent line targets physical optimization directly, using LLMs
as \textbf{frozen} components inside a search or multi-agent loop:
EvolVE \citep{evolve2025} (evolutionary code generation), VeriAgent
\citep{veriagent2025} (multi-agent system with synthesis-feedback
critic), REvolution \citep{revolution2025} (population-based search
with reflection, current SOTA), and COEVO \citep{ping2026coevo}
(co-evolution of correctness and PPA).  These four are the headline
baselines for our RTLLM~v2.0 comparison; in all cases \textbf{LLM
weights are never updated}.

\paragraph{Training-time RL on RTL.}
ChipSeek \citep{chen2025chipseek} is the closest RL-based prior work:
it trains RTL-generation models with hierarchical EDA feedback and
reports 84.09\% Pass@5 on RTLLM~v2.0, with average normalized EDAP
reduced to 0.76 in its best configuration.  EARL \citep{earl2025} is a
concurrent training-time RL method that combines SFT with
entropy-aware DAPO-style updates on verifiable compiler/testbench
signals.  Both produce an \textbf{offline, amortized policy} shared
across all designs and are evaluated under a Pass@$k$ generation
protocol.

\paragraph{Differentiation.}
Unlike functional-only Verilog generators, frozen-LLM PPA agents whose
weights never update, and offline RL methods that ship a single
amortized policy, TTT-RTL combines \emph{EDA-grounded reward} with
\emph{per-design test-time adaptation}: each problem triggers its own
on-the-fly policy update driven by a PUCT state pool, with a verifier
gated by mixed discrete--continuous physical metrics rather than
scalar pass/fail, and we report coverage and reference-normalized
PPA-product over the full RTLLM~v2.0 benchmark rather than Pass@$k$.

\section{Method: TTT-RTL}
\label{sec:method}

\begin{figure}[t]
  \centering
  \includegraphics[width=\linewidth]{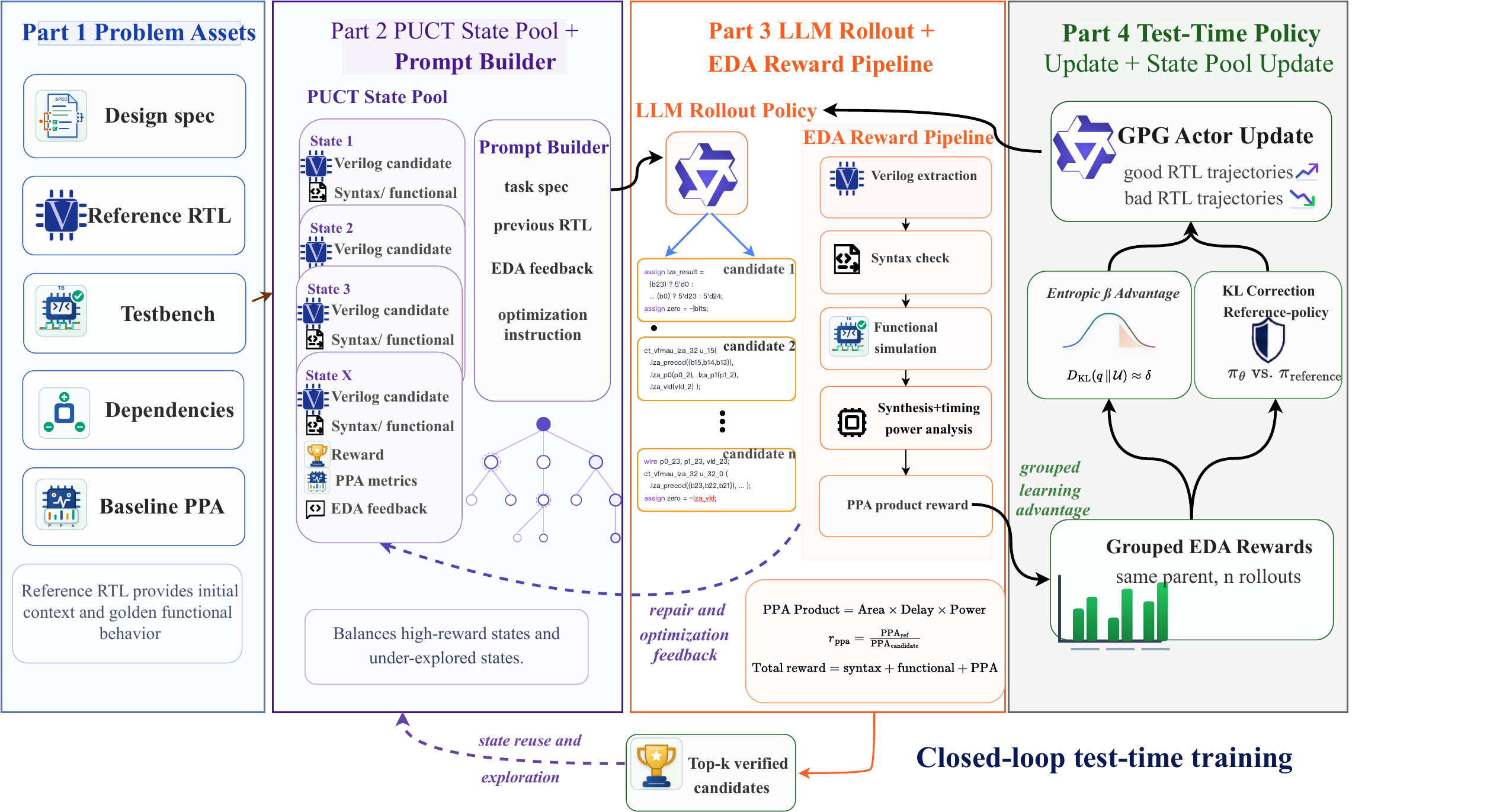}
  \caption{%
    Overview of TTT-RTL.  The framework comprises four parts:
    Problem Assets (\cref{sec:puct}), the PUCT State Pool with
    EDA-feedback prompt builder (\cref{sec:prompt}), the LLM rollout
    and three-stage EDA reward pipeline (\cref{sec:reward}), and the
    test-time policy update with state-pool admission
    (\cref{sec:advantage}).}
  \label{fig:framework}
\end{figure}

\Cref{fig:framework} summarizes the four parts of the closed-loop
TTT-RTL pipeline; the rest of this section formalizes each component.

\subsection{Problem Formulation and PUCT-Guided State Sampling}
\label{sec:puct}

Let $\mathcal{P}$ denote an RTL design problem specified by a natural-language
description, a functional testbench, and a reference implementation.
A \emph{design state} $s = (v, r, t)$ consists of a Verilog implementation $v$,
its reward $r$, and creation timestep $t$.
The goal is to find a design $v^*$ that is functionally correct and minimizes
a target metric $M(v)$ defined in \cref{sec:reward}.
We initialize a \emph{state pool} $\cS$ with a single root state
$s_0 = (\epsilon,\, 0,\, -1)$ ($\epsilon$ is an empty implementation), and grow
$\cS$ iteratively by sampling parents under a PUCT score that balances
exploitation (high-reward nodes) and exploration (rarely-sampled nodes):
\begin{equation}
  \puct(s) = \Qval(s) + c \cdot \sigma \cdot P(s) \cdot \frac{\sqrt{1 + T}}{1 + \Nvis(s)},
  \label{eq:puct}
\end{equation}
where $c = 1.0$ is the exploration coefficient, $T$ is the total number
of expanded parents so far, and $\Nvis(s)$ counts how often $s$ (or any
of its descendants) has been expanded.
$\Qval(s)$ is the best one-step reachable child reward of $s$ for
visited states, falling back to $R(s)$ when $s$ has not yet been
expanded.
The scale factor $\sigma = R_{\max} - R_{\min}$ is the reward range
taken globally over the current pool $\cS$, and $P(s)$ is the
normalized linear-rank prior
\begin{equation}
  P(s) = \frac{|\cS| - \rank(s)}{\sum_{s' \in \cS}\bigl(|\cS| - \rank(s')\bigr)},
  \qquad \rank(s) \in \{0, \ldots, |\cS|-1\},
  \label{eq:puct-prior}
\end{equation}
with states sorted in descending reward order so that $\rank(s) = 0$ is
the best state \citep[App.~A.2]{yuksekgonul2026discover}.
To maintain batch diversity, TTT-RTL applies \emph{lineage blocking}: no
two states from the same direct parent-child chain appear in one batch.

\subsection{EDA-Feedback Prompt Construction}
\label{sec:prompt}

The prompt for each parent state $s$ contains the reference implementation
with its synthesis metrics, the design specification, and---for non-root
states---the parent's Verilog and its EDA feedback (root prompts elide the
parent block):
\begin{quote}\small
\texttt{[Reference] area=\{A\_ref\}, delay=\{D\_ref\}, power=\{P\_ref\}, $M$=\{M\_ref\}.}\\
\texttt{[Spec] \{design\_description\}}\\
\texttt{[Previous] \{verilog\_code\}}\\
\texttt{[Feedback] Syntax: PASS. Functional: PASS. Synthesis: area=\{A\}, delay=\{D\}, power=\{P\}, $M$=\{M\} (improved by \{$\Delta$\}).}\\
\texttt{Produce a Verilog module with $M$ lower than \{M\}.}
\end{quote}
Stating the target as a concrete numerical threshold grounds the
generation objective in measurable physical quantities.

\subsection{Three-Stage Reward Function}
\label{sec:reward}

Each implementation $v$ is evaluated through three stages, with the reward
\begin{equation}
  r = \omega_{\text{syn}} r_{\text{syn}} + \omega_{\text{func}} r_{\text{func}} + \omega_{\text{ppa}} r_{\text{ppa}},
  \quad (\omega_{\text{syn}}, \omega_{\text{func}}, \omega_{\text{ppa}}) = (0.1, 1.0, 10.0),
  \label{eq:reward}
\end{equation}
and evaluation terminating early on failure.
\textbf{Stage 1 (syntax)}: \texttt{iverilog} compilation; on failure
$r_{\text{syn}} \in (0, 1]$ decays with the number of error messages
(detailed scoring in \cref{app:beta}).
\textbf{Stage 2 (functional)}: simulation against the reference testbench;
$r_{\text{func}} \in \{0, 1\}$.
\textbf{Stage 3 (physical synthesis)}: Yosys + OpenSTA report area $A$
($\mu\mathrm{m}^2$), critical-path delay $D$ (ps) and (where collected)
power $P$ ($\mu$W); the PPA reward is the reference-normalized ratio
\begin{equation}
  r_{\text{ppa}} = \frac{M_{\text{ref}}}{M(v)},\qquad
  M(v) = \begin{cases}
    A \cdot D \cdot P & \text{(RTLLM~v2.0 PPA product, } \ppap\text{)}\\
    A \cdot D         & \text{(C910 LZA ablation, } \adp\text{; no power)}
  \end{cases}
  \label{eq:ppa}
\end{equation}
so that $r_{\text{ppa}} = 1$ matches the reference and $r_{\text{ppa}} > 1$
strictly improves it.
The large $\omega_{\text{ppa}}$ reflects that PPA optimization is the primary
objective once functional correctness is achieved.
After scoring, the top-$k$ deduplicated children of each parent enter
$\cS$, PUCT statistics are updated, and the pool is pruned to
$C_{\max}$ if exceeded (values in \cref{tab:hyperparams}).

\subsection{Entropic Advantage Estimation}
\label{sec:advantage}

Following TTT-Discover \citep{yuksekgonul2026discover}, we estimate
advantages within each \emph{group} (rollouts sharing a common parent).
For a group of rewards $\{r_1, \ldots, r_k\}$, if
$\max r_i - \min r_i < 10^{-12}$ the group is degenerate and all
advantages are zero; otherwise we find $\betastar$ such that the softmax
distribution $q_i \propto \exp(\betastar r_i)$ attains a target KL
divergence (\emph{KL budget}) $\delta$ from uniform:
\begin{equation}
  \KL(q \| \mathcal{U}) = \sum_i q_i \ln(k \cdot q_i) = \delta,
  \label{eq:kl}
\end{equation}
solved by binary search.  Small $\delta$ keeps $q$ near uniform (diffuse
advantages); large $\delta$ allows $q$ to peak on a single rollout.
The leave-one-out advantage is
\begin{equation}
  \adv_i = \frac{\exp(\betastar r_i)}{\frac{1}{k-1}\sum_{j \neq i} \exp(\betastar r_j)} - 1,
  \label{eq:loo}
\end{equation}
which compares each rollout's exponential reward weight against the average weight of the other rollouts in the same group, yielding a group-relative contrast while avoiding self-normalization.
For RTLLM~v2.0 we use the canonical fixed $\delta = \ln 2$;
\cref{sec:adaptive-delta-variant} introduces an adaptive variant we
study on the C910 LZA case study (\cref{sec:c910_kl_budget}).

\subsection{Policy Gradient Update}

For rollout $i$ with prompt $x_i$, the policy loss is
\begin{equation}
  \mathcal{L}(\vtheta) = -\log \piTheta(v_i | x_i) \cdot \adv_i,
  \label{eq:loss}
\end{equation}
averaged over non-degenerate rollouts; following TTT-Discover, we omit
the PPO importance-ratio clip \citep{schulman2017ppo}.

\subsection{Full Algorithm}

\begin{algorithm}
\caption{TTT-RTL}
\label{alg:ttt-rtl}
\begin{algorithmic}[1]
  \REQUIRE Design problem $\mathcal{P}$ (spec, testbench, reference),
           LLM $\piTheta$, budget $S$, batch size $B$, rollouts per prompt $n$
  \STATE Initialize pool $\cS \leftarrow \{s_0\}$ (root), $T \leftarrow 0$
  \FOR{step $= 1$ \TO $S$}
    \STATE Sample $B$ parent states via \cref{eq:puct} with lineage blocking
    \STATE Construct prompts $\{x_b\}_{b=1}^{B}$ (\cref{sec:prompt})
    \STATE Generate $n$ rollouts per prompt using $\piTheta$ via vLLM \citep{kwon2023vllm}
    \STATE Evaluate rewards via 3-stage EDA pipeline (\cref{sec:reward})
    \STATE Update pool $\cS$, PUCT statistics, prune to $C_{\max}$
    \STATE Compute advantages via entropic $\betastar$ at fixed $\delta = \ln 2$ (\cref{sec:advantage})
    \STATE Update $\vtheta$ via policy gradient \cref{eq:loss}
  \ENDFOR
  \RETURN Best-reward design in $\cS$
\end{algorithmic}
\end{algorithm}

\subsection{Adaptive KL-budget controller}
\label{sec:adaptive-delta-variant}

For RTLLM~v2.0 we use the canonical fixed $\delta = \ln 2$.
On the C910 LZA case study (\cref{sec:c910_kl_budget}) we additionally
study an adaptive variant that replaces the constant with $\delta_t$
updated once per step from four EMA-smoothed signals---policy-vs-reference
KL, effective number of distinct rollouts per group, fraction of
constant-reward groups, and binary-search saturation rate---combined
through a four-rule priority ladder (KL brake, winner-take-all,
stagnation, over-exploring) that resets, shrinks, or grows $\delta_t$
within $[0.25\ln 2,\, 4\ln 2]$.  Full equations
(\cref{eq:adaptive-ema}--\cref{eq:p4}) and hyperparameters
(\cref{tab:adaptive_hparams}) are in \cref{app:beta}.

\section{Experiments}
\label{sec:experiments}

\subsection{Experimental Setup}

\paragraph{Base model and training configuration.}
The policy model is Qwen3-8B \citep{yang2025qwen3}.  RTLLM~v2.0 main
runs initialize from a lightweight format-and-style SFT warm-up
(\texttt{ttt-rtl-sft} step 18; see \cref{app:sft}); the C910 LZA
ablations (\cref{sec:ablations}) instead use the \emph{raw} Qwen3-8B
base model with no SFT, isolating the contribution of test-time RL
on top of an off-the-shelf backbone.  Each run is 100 gradient steps
with $B = 4$ parent states, top-$k = 2$ children, exploration
coefficient $c = 1.0$, and the reward weights of \cref{eq:reward};
RTLLM~v2.0 uses $n = 4$ rollouts/prompt and C910 LZA ablations use
$n = 8$, matching TTT-Discover's per-step budget
\citep{yuksekgonul2026discover}.
RTLLM~v2.0 uses the canonical fixed $\delta = \ln 2$; C910 ablations
vary $\delta$ along one axis of \cref{tab:ablations}.  Full
hyperparameters are in \cref{tab:hyperparams}.

\paragraph{Benchmark and baselines.}
RTLLM~v2.0 \citep{lu2024rtllm, liu2025openllmrtl} provides a
human-written reference and testbench for each of 49 designs.  We
report the PPA-product ratio $\ppap = A \cdot D \cdot P$ of each
method's best functionally correct design over the v2.0 reference
(lower is better; $1.0$ matches the reference).  Baselines are three
published agent-based methods that target the same metric:
\textbf{EvolVE} \citep{evolve2025}, \textbf{VeriAgent}
\citep{veriagent2025}, and the current SOTA \textbf{REvolution}
\citep{revolution2025}; per-design ratios are taken verbatim from
Table~3 of \citet{ping2026coevo}, where every baseline runs under
GPT-4o-mini against the v2.0 reference under the same PPA-product
metric.

\paragraph{Synthesis flow.}
Yosys \citep{wolf2013yosys} for logic synthesis, OpenSTA for
timing/power.  RTLLM~v2.0 uses Nangate~45\,nm typical-corner (matching
the published baseline flow); C910 LZA uses Sky130 HD (the C910
release convention).  All comparisons are reference-normalized within
a single PDK; we discuss residual cross-flow uncertainty in
\cref{app:flow_sanity}.

\paragraph{Compute and seed protocol.}
All runs use A800 GPUs and seed $42$, in line with the baselines
(none of EvolVE / VeriAgent / REvolution report seed variance); a
four-seed paired replication on LZA \texttt{simd\_half} and a
single-seed case study on a second C910 unit are reported in
\cref{app:kl_budget_multiseed}.

\subsection{RTLLM v2.0: Framework vs Published Agent Baselines}
\label{sec:main_results}

\Cref{tab:main_results} aggregates the per-design PPA-product ratios across all
$49$ RTLLM v2.0 problems; the full per-design table, including the $5$
designs where at least one method failed to produce compilable,
functionally correct RTL, is in \cref{tab:main_full} (\cref{app:main_full}).
We characterize the comparison along three complementary views, all of
which point in the same direction.
\textbf{(i) Coverage:} TTT-RTL produces a valid implementation on
$48/49$ designs (highest of any method evaluated; missing only
\texttt{asyn\_fifo}), and is the only method that succeeds on
\texttt{freq\_divbyfrac} and \texttt{freq\_divbyodd}.
\textbf{(ii) Intersection GeoMean ($N{=}44$):} on the subset where
all four methods produced a valid output --- the only domain on which
GeoMean is mathematically well-defined for direct comparison ---
TTT-RTL achieves a geometric-mean PPA-product ratio of $0.349$ versus
$0.739$ for the strongest published baseline (REvolution).
\textbf{(iii) Penalized full-benchmark GeoMean ($N{=}49$,
failures$=1.0$):} imputing each method's failures as ratio $1.0$
(``no improvement over the reference'', the most lenient possible
imputation since it credits a failed method with matching the
reference) and recomputing GeoMean over the full $49$ designs gives
$0.341$ for TTT-RTL versus $0.762$ for REvolution
(\cref{tab:main_results}, last column).
The intersection metric ($N{=}44$) is therefore not a cherry-pick: any
defensible way of folding baseline failures into the comparison only
widens the TTT-RTL margin, never narrows it.

\begin{table}[t]
  \centering
  \caption{%
    Reference-normalized external comparison on RTLLM~v2.0.  Each cell
    reports the PPA-product ratio ($\ppap = A \cdot D \cdot P$) of a
    method's best functionally correct design over the v2.0 reference
    implementation; lower is better.  Baseline ratios are taken from
    Table~3 of \citet{ping2026coevo} (GPT-4o-mini), while
    TTT-RTL is evaluated under our matched Yosys / OpenSTA /
    Nangate~45\,nm flow.  \emph{Coverage} counts designs (out of $49$)
    on which a method produced compilable, functionally correct RTL;
    \emph{GeoMean / ArithMean$_{N{=}44}$} are over the common subset
    where all four methods completed; \emph{\#Improved} / \emph{\#Best}
    count common-subset designs with ratio $<1$ / lowest ratio (ties
    split equally); \emph{GeoMean$_{N{=}49}^{\mathrm{pen}}$} imputes
    each method's failures as ratio $1.0$ (subset choices and
    robustness check are detailed below).  See \cref{sec:main_results}
    and \cref{app:flow_sanity} for the flow sanity check.
  }
  \label{tab:main_results}
  \small
  \begin{tabular}{lccccccc}
    \toprule
    Method & Coverage & GeoMean$_{N{=}44}\downarrow$ & ArithMean$_{N{=}44}\downarrow$ & \#Improved & \#Best & GeoMean$_{N{=}49}^{\mathrm{pen}}\downarrow$ \\
    \midrule
    EvolVE     & 46/49 & 0.872 & 0.909 & 17/44 &  0.5/44 & 0.892 \\
    VeriAgent  & 44/49 & 0.813 & 0.853 & 25/44 &  2.0/44 & 0.830 \\
    REvolution & 45/49 & 0.739 & 0.813 & 29/44 &  8.0/44 & 0.762 \\
    \textbf{TTT-RTL}
               & \textbf{48/49} & \textbf{0.349} & \textbf{0.527}
               & \textbf{38/44} & \textbf{33.5/44} & \textbf{0.341} \\
    \bottomrule
  \end{tabular}
\end{table}

\begin{figure}[t]
  \centering
  \includegraphics[width=\linewidth]{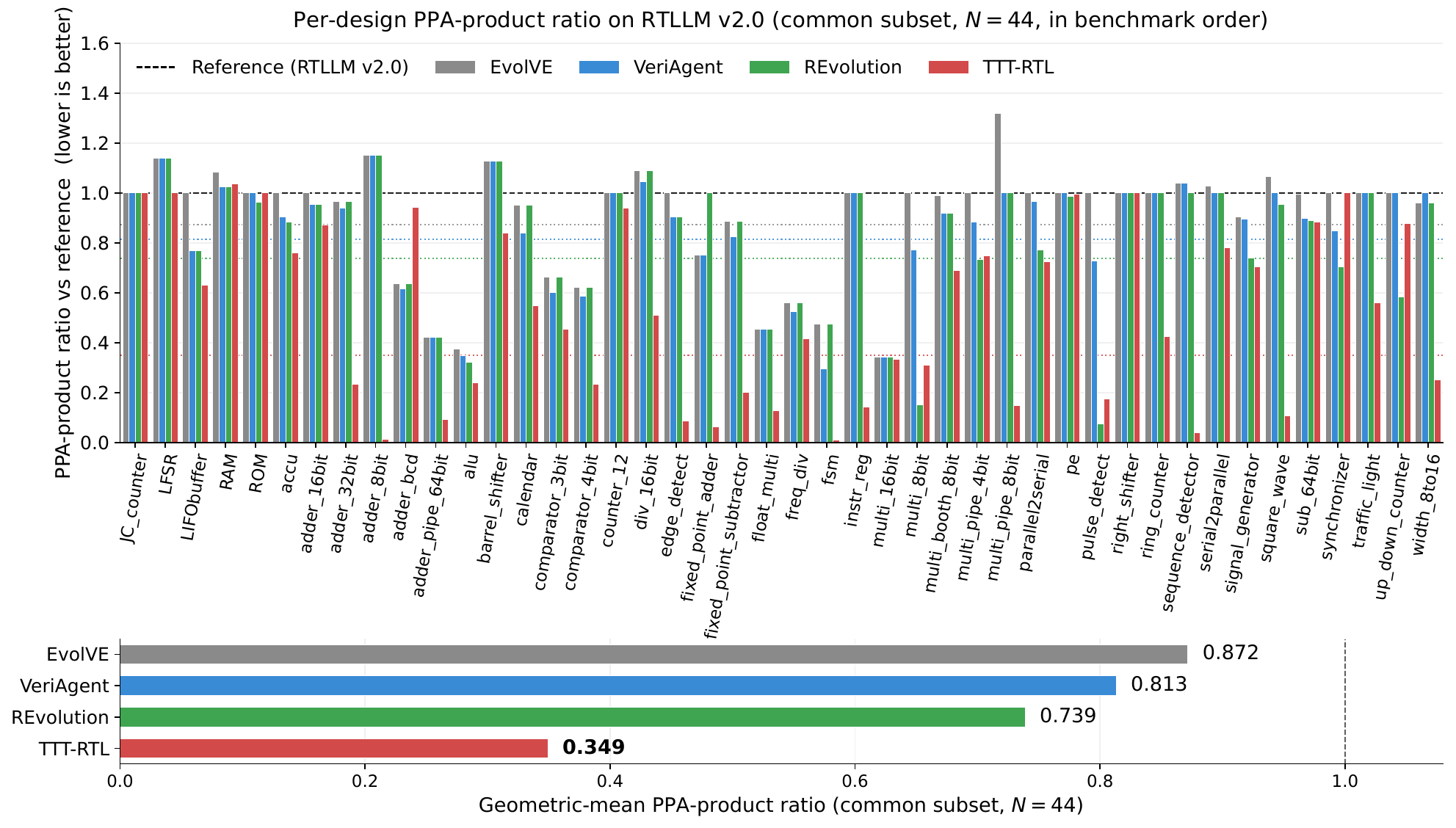}
  \caption{%
    Per-design PPA-product ratio ($\ppap = A \cdot D \cdot P$) on RTLLM v2.0
    (the $N{=}44$ common subset, listed in benchmark order; the $5$
    designs with at least one missing method are shown in
    \cref{tab:main_full}).
    The EvolVE / VeriAgent / REvolution per-design ratios are the
    GPT-4o-mini numbers reported in Table~3 of
    \citet{ping2026coevo}; TTT-RTL ratios are produced under the
    flow described in \cref{sec:experiments}.
    \textbf{Top:} four-method bar chart; the dashed black line is the v2.0
    reference ($1.0$), and the colored dotted lines are each method's
    $N{=}44$ geometric-mean ratio.
    \textbf{Bottom:} aggregate geometric-mean PPA-product ratio per method
    on the $N{=}44$ common subset.
    TTT-RTL is below the reference on $38/44$ designs and is the
    best method on $33.5/44$ (next-best is REvolution at $8/44$).
  }
  \label{fig:main_results}
\end{figure}

The advantage is consistent across complexity bins
(\cref{tab:main_by_complexity}), including on small combinational
designs where prior methods barely improve.  Full per-design ratios
and the failure breakdown are in \cref{tab:main_full,app:main_full}.

\begin{table}[t]
  \centering
  \caption{%
    Geometric-mean PPA-product ratio on RTLLM v2.0, broken down by design complexity
    (binned by reference PPA product).
    Lower is better; bold marks the best method per row.
  }
  \label{tab:main_by_complexity}
  \small
  \begin{tabular}{lrcccc}
    \toprule
    Complexity bin & $N$ & EvolVE & VeriAgent & REvolution & TTT-RTL \\
    \midrule
    Small (ref PPA $<10^{3}$)            & 17 & 0.902 & 0.836 & 0.731 & \textbf{0.305} \\
    Medium ($10^{3}$--$10^{5}$)          & 14 & 0.963 & 0.906 & 0.901 & \textbf{0.475} \\
    Large ($10^{5}$--$10^{7}$)           & 10 & 0.736 & 0.675 & 0.559 & \textbf{0.274} \\
    Huge ($>10^{7}$)                     &  3 & 0.790 & 0.779 & 0.787 & \textbf{0.403} \\
    \bottomrule
  \end{tabular}
\end{table}

Of the five non-intersection designs, four are ones our flow handles
where one or more baselines failed, so excluding them is conservative
for TTT-RTL rather than a cherry-pick; per-design failures are listed
in \cref{app:main_full}.  A small number of v2.0 designs hit the
OpenSTA $0$\,ps delay-reporting floor (PPA product collapses to
area-only), which affects all four methods uniformly under the shared
reference; we flag this as a reward-shaping limitation in the
NeurIPS checklist (Limitations).

\paragraph{Caveat: external systems comparison, not same-backbone isolation.}
The four methods in \cref{tab:main_results} share the same v2.0
reference, the same PPA-product metric, and (for our re-runs) the
same Yosys$+$OpenSTA flow, so the comparison is reference-normalized
and ratio-controlled.  The methods do, however, use different policy
LLMs: EvolVE, VeriAgent, and REvolution use GPT-4o-mini (as reported
in \citet{ping2026coevo}, Table~3), while TTT-RTL uses Qwen3-8B with
domain SFT plus per-design test-time training.  We therefore read
\cref{tab:main_results} as a \emph{systems} comparison---per-design
test-time training on top of an open-source backbone vs.\ frozen
agentic search on top of a stronger commercial backbone---rather than
as a pure algorithmic isolation.  A frozen Best-of-$N$ baseline on
the C910 unit (\cref{tab:ablations}, ``Best-of-$N$'' row), which uses
the raw Qwen3-8B base model with no SFT and no test-time updates,
never produces a functionally correct design within the $3200$-rollout
budget, supporting that test-time updates---not the backbone or the
SFT warm-up---drive the framework's gains.

\subsection{XuanTie C910 LZA: Industrial Case Study}
\label{sec:ablations}
\label{sec:c910_components}
\label{sec:c910_kl_budget}

The RTLLM~v2.0 results above establish that TTT-RTL beats published
agent baselines on a textbook benchmark.  To stress-test the framework
on \emph{production silicon}, we now turn to the leading-zero-anticipation
unit \texttt{ct\_vfmau\_lza\_simd\_half} from the open-source XuanTie
C910 RISC-V core \citep{chen2020xuantie}---well-tuned hand-written RTL
where any ADP reduction is non-trivial.  We use this single industrial
unit to ablate TTT-RTL along three axes: the reuse strategy, the
training-time advantage estimator, and the KL-budget schedule.
All runs share the same sampling budget (100 steps $\times$ 4 parents
$\times$ $n=8$ rollouts $= 3200$ rollouts), base model (Qwen3-8B),
technology library (Sky130), seed ($42$), and reward function
(\cref{eq:reward}).  The reference ADP for this unit is
$3.40$M\,$\mu\mathrm{m}^2{\cdot}\mathrm{ps}$, at which the reward is
exactly~$11.1$; any reward above $11.1$ corresponds to a functionally
correct design that strictly improves ADP.

\paragraph{Ablation axes.}
We sweep three orthogonal axes one at a time.  On the
\emph{$\delta$-schedule} axis, the other two axes are fixed at the
full TTT-RTL configuration (PUCT reuse, entropic advantage), and the
``adaptive $\delta$'' row \emph{is} the headline full-TTT-RTL row at
the top of \cref{tab:ablations}.  On the \emph{reuse} and \emph{train}
axes we instead hold $\delta$ at the fixed $\ln 2$ TTT-Discover
default rather than at adaptive $\delta$, so that the
component-isolation comparisons do not implicitly inherit the
controller's exploratory gain; the framework-only ($-45.3\%$) row is
the natural reference for those rows.
The three axes are then
(i) \textbf{reuse} $\in$ \{PUCT, $\epsilon$-greedy ($\epsilon{=}0.1$),
none\}, where ``none'' restarts every rollout from the empty root;
(ii) \textbf{train} $\in$ \{entropic advantage, expected reward\},
where the latter is standard GRPO \citep{shao2024deepseekmath} with
group-mean-centred, std-normalised advantages and no entropic
temperature;
(iii) \textbf{$\delta$-schedule} $\in$ \{adaptive (controller of
\cref{sec:adaptive-delta-variant}), fixed $\ln 2$ (TTT-Discover's
constant), cosine $1.1\!\to\!0.3$ (``explore-then-exploit''), cosine
$0.3\!\to\!1.1$ (reverse, sanity-check)\}.
Two combined-weakest configurations bound the ablation from below:
\emph{Naive Test-time RL} (expected reward + no reuse) and
\emph{Best-of-$N$} (frozen actor at lr $10^{-12}$, $N{=}3200$).

\begin{table}[t]
  \centering
  \caption{%
    Ablation results on \texttt{ct\_vfmau\_lza\_simd\_half} (Sky130,
    Qwen3-8B, seed $42$, $3200$-rollout budget).
    \textbf{All rows are single-seed.}
    Reward $11.1$ corresponds to the reference ADP; entries marked
    ``---'' never produced any functionally correct design within the
    budget.  Per-step reward trajectories are in \cref{fig:ablations};
    KL-budget trajectories are in \cref{app:kl_budget_trajectories}.
    Quantitative gaps within the table should be read as consistent
    ranking evidence rather than tight effect sizes.
  }
  \label{tab:ablations}
  \footnotesize
  \setlength{\tabcolsep}{5pt}
  \begin{tabular}{lrrr}
    \toprule
    \textbf{Configuration}
       & \textbf{Best reward ($\uparrow$)}
       & \textbf{Best ADP ($\mu\mathrm{m}^2{\cdot}\mathrm{ps}$, $\downarrow$)}
       & \textbf{ADP reduction} \\
    \midrule
    Reference (C910 baseline)
       & $11.10$ & $3{,}402{,}207$ & --- \\
    \midrule
    \textbf{TTT-RTL (PUCT + entropic + adaptive $\delta$)}
       & $\mathbf{25.72}$ & $\mathbf{1{,}381{,}970}$ & $\mathbf{-59.4\%}$ \\
    \midrule
    \multicolumn{4}{l}{\emph{$\delta$-schedule axis (reuse = PUCT, train = entropic)}} \\
    \quad cosine $1.1\!\to\!0.3$ (high$\to$low)
       & $21.02$ & $1{,}708{,}247$ & $-49.8\%$ \\
    \quad fixed $\delta = \ln 2$ (TTT-Discover default)
       & $19.38$ & $1{,}860{,}973$ & $-45.3\%$ \\
    \quad cosine $0.3\!\to\!1.1$ (low$\to$high)
       & $15.91$ & $2{,}296{,}653$ & $-32.5\%$ \\
    \midrule
    \multicolumn{4}{l}{\emph{Train axis (reuse = PUCT, $\delta$ = fixed $\ln 2$)}} \\
    \quad expected reward (no entropic)
       & $16.65$ & $2{,}187{,}460$ & $-35.7\%$ \\
    \midrule
    \multicolumn{4}{l}{\emph{Reuse axis (train = entropic, $\delta$ = fixed $\ln 2$)}} \\
    \quad $\epsilon$-greedy ($\epsilon{=}0.1$)
       & $21.69$ & $1{,}652{,}580$ & $-51.4\%$ \\
    \quad no reuse
       & $12.63$ & $2{,}951{,}048$ & $-13.3\%$ \\
    \midrule
    Naive Test-time RL (expected reward + no reuse)
       & $0.89$  & --- & --- \\
    Best-of-$N$ (frozen policy + no reuse, $N{=}3200$)
       & $1.00$  & --- & --- \\
    \bottomrule
  \end{tabular}
\end{table}

\begin{figure}[t]
  \centering
  \includegraphics[width=\linewidth]{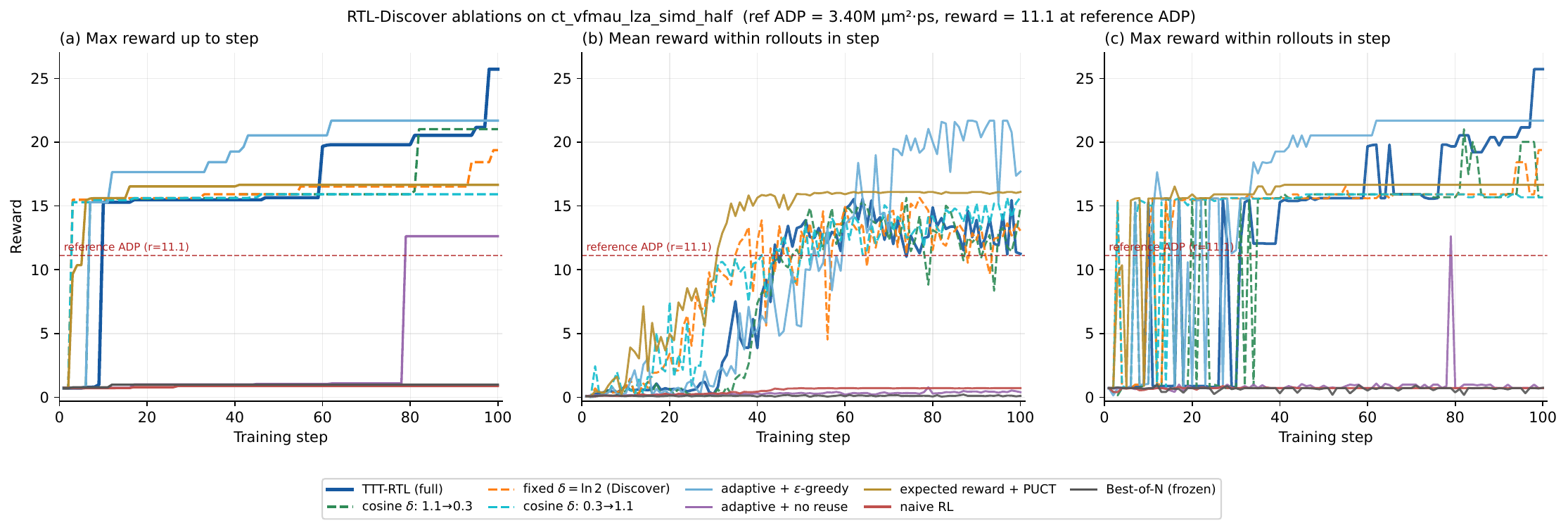}
  \caption{%
    Ablation trajectories on \texttt{ct\_vfmau\_lza\_simd\_half},
    following the three-panel format of
    \citet{yuksekgonul2026discover}, Fig.~10.
    \textbf{(a) Max reward up to step}; \textbf{(b) Mean reward within
    rollouts in step} (typical policy behaviour); \textbf{(c) Max
    reward within rollouts in step} (per-step exploration quality).
    The dashed red line marks $r = 11.1$ (reference ADP); designs
    above the line strictly improve over the C910 baseline.
    All runs share the same $3200$-rollout budget, base model, reward,
    and seed; the only differences are the train/reuse components as
    in \cref{tab:ablations}.
  }
  \label{fig:ablations}
\end{figure}

\paragraph{Reuse and train axes.}
With $\delta$ held at the TTT-Discover default $\ln 2$, the framework
alone (PUCT + entropic + fixed $\delta$) reaches $-45.3\%$ ADP.
Removing the entropic advantage estimator drops this to $-35.7\%$
(plateauing after step $\sim\!30$), while replacing PUCT with
$\epsilon$-greedy reuse closes most of the reuse-side gap
($-51.4\%$ at seed 42).  Removing reuse entirely is the most costly
single-axis change: $-13.3\%$.  Naive-RL (expected reward + no reuse)
and Best-of-$N$ on the frozen policy never produce a single
functionally correct design within the $3200$-rollout budget.  The
reuse-vs-$\epsilon$-greedy gap is small enough that we read it as
``PUCT helps but is not the sole driver''; the entropic-vs-expected
gap is larger and consistent with \citet{yuksekgonul2026discover}'s
observation that exploration matters most outside kernel-engineering
settings, where uniform sampling from the base policy almost never
yields a correct Verilog module.

\paragraph{$\delta$-schedule axis.}
On top of the framework's $-45.3\%$, the adaptive $\delta$-controller
adds another $\sim\!14$ pp at seed $42$ ($-59.4\%$).  The cosine
$1.1\!\to\!0.3$ schedule, hand-tuned to approximate ``explore early,
exploit late,'' recovers most of the gap ($-49.8\%$); the reverse
cosine is uniformly worse.  This is consistent with the controller
firing P4 (over-exploring, grow $\delta$) when advantages are diffuse
early on, and P2/P3 (winner-take-all / stagnation, shrink $\delta$)
once a winner emerges---i.e.\ the trajectory resembles the
explore-then-exploit shape that the cosine encodes by hand.
Per-step KL-budget trajectories appear in \cref{app:kl_budget_trajectories}.
A four-seed paired replication on LZA \texttt{simd\_half} and a
single-seed case study on a second C910 unit
(\cref{app:kl_budget_multiseed}) confirm task-responsive direction
and a ${\sim}2.6\times$ reduction in seed-wise variance, but the gap
does not reach $p < 0.05$ at $n = 4$; we discuss this caveat further
in \cref{app:kl_budget_multiseed}.

\paragraph{Mean vs.\ max reward.}
\Cref{fig:ablations}b,c isolates the entropic advantage estimator: expected-reward + PUCT lifts the per-step \emph{mean} reward but caps the per-step \emph{max} below TTT-RTL's, trading a slightly worse average rollout for a consistently better best rollout---the quantity that matters in a discovery setting.

\section{Conclusion}
\label{sec:conclusion}

We presented TTT-RTL, a framework that closes the loop between an LLM
policy and EDA synthesis tools to apply test-time training to RTL
optimization.  On RTLLM~v2.0 it cuts the PPA product against the
strongest published agent baseline, and on a XuanTie~C910 industrial
LZA unit it improves over well-tuned production RTL while
component ablations isolate the contribution of the PUCT state pool
and the entropic advantage estimator.

\bibliography{references}
\bibliographystyle{plainnat}

\newpage
\appendix
\section{Per-design Results on RTLLM v2.0}
\label{app:main_full}

\Cref{tab:main_full} reports the per-design ratio of every method on every
RTLLM v2.0 problem (all $49$ designs).
``--'' indicates that the method failed to produce a compilable / functionally
correct design within its reported budget.
The TTT-RTL column is the PPA-product ratio ($A \cdot D \cdot P$) that
drove its training reward; the EvolVE / VeriAgent / REvolution columns
are the PPA-product ratios reported by the original authors against
the same v2.0 reference (see ``Caveat: external systems comparison,
not same-backbone isolation'' in
\cref{sec:main_results}).
The summary statistics (\cref{tab:main_results} in the main text) are
computed over the $N{=}44$ common subset where all four methods
completed (the intersection required for a fair GeoMean); the five
remaining designs (\texttt{asyn\_fifo}, \texttt{radix2\_div},
\texttt{freq\_divbyeven}, \texttt{freq\_divbyfrac},
\texttt{freq\_divbyodd}) are excluded from GeoMean / \#Best but are
shown in the per-design table below for completeness.

\begin{table}[ht]
\centering\footnotesize
\caption{Per-design ratio on RTLLM v2.0 (lower is better; bold = best per row; ``--'' = method failed to produce compilable RTL). All four methods report the PPA-product ratio $A{\cdot}D{\cdot}P$ against the same v2.0 reference; EvolVE / VeriAgent / REvolution numbers are taken from \citet{ping2026coevo}, Table~3 (GPT-4o-mini).}
\label{tab:main_full}
\begin{tabular}{lrcccc}
\toprule
Design & ref $M$ & EvolVE & VeriAgent & REvolution & TTT-RTL \\
\midrule
JC\_counter & 10180.4 & \textbf{1.000} & \textbf{1.000} & \textbf{1.000} & \textbf{1.000} \\
LFSR & 93.5 & 1.140 & 1.140 & 1.140 & \textbf{1.000} \\
LIFObuffer & 28866.5 & 1.000 & 0.770 & 0.770 & \textbf{0.632} \\
RAM & 63406.1 & 1.083 & \textbf{1.026} & \textbf{1.026} & 1.036 \\
ROM & 46.3 & 1.000 & 1.000 & \textbf{0.964} & 1.000 \\
accu & 82040.8 & 1.000 & 0.904 & 0.885 & \textbf{0.760} \\
adder\_16bit & 3513.1 & 1.000 & 0.955 & 0.955 & \textbf{0.872} \\
adder\_32bit & 25037.3 & 0.967 & 0.939 & 0.967 & \textbf{0.235} \\
adder\_8bit & 440.0 & 1.152 & 1.152 & 1.152 & \textbf{0.014} \\
adder\_bcd & 484.6 & 0.638 & \textbf{0.617} & 0.638 & 0.943 \\
adder\_pipe\_64bit & 6109235.7 & 0.423 & 0.423 & 0.423 & \textbf{0.094} \\
alu & 3192539.6 & 0.375 & 0.348 & 0.321 & \textbf{0.241} \\
asyn\_fifo & 580752.5 & -- & -- & -- & -- \\
barrel\_shifter & 94.4 & 1.127 & 1.127 & 1.127 & \textbf{0.840} \\
calendar & 9374.5 & 0.953 & 0.840 & 0.953 & \textbf{0.548} \\
comparator\_3bit & 10.2 & 0.664 & 0.600 & 0.664 & \textbf{0.454} \\
comparator\_4bit & 26.9 & 0.621 & 0.586 & 0.621 & \textbf{0.234} \\
counter\_12 & 824.7 & 1.000 & 1.000 & 1.000 & \textbf{0.940} \\
div\_16bit & 94019995.7 & 1.089 & 1.045 & 1.089 & \textbf{0.510} \\
edge\_detect & 40.5 & 1.000 & 0.904 & 0.904 & \textbf{0.086} \\
fixed\_point\_adder & 233273.1 & 0.752 & 0.752 & 1.000 & \textbf{0.062} \\
fixed\_point\_subtractor & 354996.9 & 0.888 & 0.824 & 0.888 & \textbf{0.201} \\
float\_multi & 914343626.2 & 0.453 & 0.453 & 0.453 & \textbf{0.129} \\
freq\_div & 4986.7 & 0.561 & 0.526 & 0.561 & \textbf{0.417} \\
freq\_divbyeven & 290.5 & 1.574 & -- & -- & \textbf{0.607} \\
freq\_divbyfrac & 388.2 & -- & -- & -- & \textbf{0.561} \\
freq\_divbyodd & 2347.5 & -- & -- & -- & \textbf{0.425} \\
fsm & 429.1 & 0.476 & 0.297 & 0.476 & \textbf{0.011} \\
instr\_reg & 2415.7 & 1.000 & 1.000 & 1.000 & \textbf{0.143} \\
multi\_16bit & 2354913.0 & 0.342 & 0.342 & 0.342 & \textbf{0.335} \\
multi\_8bit & 725109.1 & 1.000 & 0.771 & \textbf{0.153} & 0.311 \\
multi\_booth\_8bit & 196035.4 & 0.989 & 0.918 & 0.918 & \textbf{0.691} \\
multi\_pipe\_4bit & 11025.1 & 1.000 & 0.884 & \textbf{0.734} & 0.750 \\
multi\_pipe\_8bit & 584498.4 & 1.319 & 1.000 & 1.000 & \textbf{0.149} \\
parallel2serial & 450.8 & 1.000 & 0.965 & 0.771 & \textbf{0.726} \\
pe & 224095126.1 & 1.000 & 1.000 & \textbf{0.986} & 0.996 \\
pulse\_detect & 153.2 & 1.000 & 0.727 & \textbf{0.075} & 0.174 \\
radix2\_div & 88514.0 & 0.973 & -- & 0.973 & \textbf{0.011} \\
right\_shifter & 108.5 & \textbf{1.000} & \textbf{1.000} & \textbf{1.000} & \textbf{1.000} \\
ring\_counter & 263.1 & 1.000 & 1.000 & 1.000 & \textbf{0.425} \\
sequence\_detector & 211.0 & 1.040 & 1.040 & 1.000 & \textbf{0.041} \\
serial2parallel & 14019.8 & 1.027 & 1.000 & 1.000 & \textbf{0.782} \\
signal\_generator & 678.1 & 0.906 & 0.894 & 0.738 & \textbf{0.705} \\
square\_wave & 12354.9 & 1.066 & 1.000 & 0.954 & \textbf{0.108} \\
sub\_64bit & 238797.5 & 0.995 & 0.899 & 0.890 & \textbf{0.885} \\
synchronizer & 671.8 & 1.000 & 0.847 & \textbf{0.706} & 1.000 \\
traffic\_light & 32088.5 & 1.000 & 1.000 & 1.000 & \textbf{0.561} \\
up\_down\_counter & 101304.7 & 1.000 & 1.000 & \textbf{0.584} & 0.878 \\
width\_8to16 & 14891.7 & 0.960 & 1.000 & 0.960 & \textbf{0.253} \\
\midrule
GeoMean (common, $N{=}44$) & -- & 0.872 & 0.813 & 0.739 & \textbf{0.349} \\
\#Best / 44 & -- & 0.5 & 2.0 & 8.0 & \textbf{33.5} \\
\bottomrule
\end{tabular}
\end{table}

\section{Flow Sanity Check: Local vs.\ Published Reference PPA}
\label{app:flow_sanity}

To support the ratio-normalized comparison protocol described in
\cref{sec:main_results} (``Flow sanity check and ratio-normalized
comparison''), we re-evaluated the RTLLM~v2.0 reference Verilog on
every design under our local Yosys$+$OpenSTA flow with the
Nangate~45\,nm typical-corner library, and compared the resulting
area / delay / power against the published reference numbers reported
in Table~3 of \citet{ping2026coevo} (the open-source pipeline
released as \texttt{hping666/COEVO}).
\Cref{tab:flow_sanity_repr} shows representative rows spanning the
four complexity bins of \cref{tab:main_by_complexity};
\cref{tab:flow_sanity_summary} reports aggregate error statistics
over the designs that synthesize cleanly under both flows ($47$ for
area, $43$ for delay/power; see the table caption).

\begin{table}[h]
  \centering
  \footnotesize
  \caption{%
    Representative per-design comparison of the v2.0 reference RTL
    under our local flow vs.\ the published numbers
    \citep{ping2026coevo}.  Area in $\mu\mathrm{m}^2$, delay in
    ns, power in $\mu$W; the $\Delta$ columns are signed relative
    error $(\text{ours} - \text{pub}) / \text{pub}$.  Area and delay
    are closely matched on the vast majority of designs; absolute
    power shows larger discrepancies driven by OpenSTA activity /
    reporting assumptions, which is the reason we only report ratios
    against the shared reference in the main comparison.  Two
    outlier-style cases (\texttt{RAM} and \texttt{asyn\_fifo}, where
    Yosys infers different array / synchronizer structures across
    flows) are included for completeness.
  }
  \label{tab:flow_sanity_repr}
  \begin{tabular}{lrrrrrrrrr}
    \toprule
    Design & $A_{\text{pub}}$ & $A_{\text{ours}}$ & $\Delta A$
           & $D_{\text{pub}}$ & $D_{\text{ours}}$ & $\Delta D$
           & $P_{\text{pub}}$ & $P_{\text{ours}}$ & $\Delta P$ \\
    \midrule
    comparator\_3bit  &   12.0 &   12.0 & $+0.0\%$ & 0.15 & 0.16 &  $+6.7\%$ &    5.7 &   4.7 & $-18.3\%$ \\
    adder\_8bit       &   48.9 &   48.9 & $+0.0\%$ & 0.31 & 0.34 &  $+9.7\%$ &   29.0 &  16.8 & $-42.1\%$ \\
    calendar          &  164.1 &  164.9 & $+0.5\%$ & 0.42 & 0.47 & $+11.9\%$ &  136.0 & 105.0 & $-22.8\%$ \\
    adder\_16bit      &   96.8 &   94.7 & $-2.2\%$ & 0.59 & 0.67 & $+13.6\%$ &   61.5 &  40.4 & $-34.3\%$ \\
    adder\_32bit      &  208.5 &  211.2 & $+1.3\%$ & 0.87 & 0.91 &  $+4.6\%$ &  138.0 &  85.1 & $-38.3\%$ \\
    alu               & 1953.0 & 1967.1 & $+0.7\%$ & 1.93 & 2.10 &  $+8.8\%$ &  847.0 & 688.0 & $-18.8\%$ \\
    multi\_16bit      &  951.5 &  933.9 & $-1.8\%$ & 1.98 & 2.19 & $+10.6\%$ & 1250.0 & 796.0 & $-36.3\%$ \\
    adder\_pipe\_64bit & 2529.4 & 2523.5 & $-0.2\%$ & 0.83 & 0.90 &  $+8.4\%$ & 2910.0 & 1920.0 & $-34.0\%$ \\
    pe                & 3649.5 & 3640.2 & $-0.3\%$ & 1.72 & 1.91 & $+11.0\%$ & 35700.0 & 1400.0 & $-96.1\%$ \\
    \midrule
    \multicolumn{10}{l}{\emph{Outlier-style cases (different inferred structure across flows)}} \\
    RAM               &  474.8 &  633.9 & $+33.5\%$ & 0.22 & 0.32 & $+45.5\%$ &  607.0 & 479.0 & $-21.1\%$ \\
    asyn\_fifo        & 1099.9 & 1341.7 & $+22.0\%$ & 0.33 & 0.87 & $+163.6\%$ & 1600.0 & 1140.0 & $-28.7\%$ \\
    \bottomrule
  \end{tabular}
\end{table}

\begin{table}[h]
  \centering
  \footnotesize
  \caption{%
    Aggregate flow-sanity statistics over the RTLLM~v2.0 designs that
    synthesize cleanly under both flows.  The \emph{Area} row covers
    all $47$ designs that complete area synthesis under both flows;
    the \emph{Delay} and \emph{Power} rows are restricted to the
    $43$-design subset for which both flows also produce a valid
    OpenSTA timing report (designs where one flow returns a
    delay-floor or missing-power report are excluded from those two
    rows only).  Error is signed relative
    error of our local measurement against the published value;
    \emph{within-$X\%$} counts designs whose absolute relative error
    is at most $X$.  The takeaway is that the PPA product's area and
    delay components are tightly matched, and the residual cross-flow
    discrepancy lives almost entirely in absolute power---which the
    main-result protocol absorbs by reporting only ratios against the
    shared v2.0 reference.
  }
  \label{tab:flow_sanity_summary}
  \begin{tabular}{lcccc}
    \toprule
    Component & median $|{\rm err}|$ & mean $|{\rm err}|$ & within $5\%$ & within $15\%$ \\
    \midrule
    Area  & $0.6\%$  & $6.9\%$  & $36/47$ & $43/47$ \\
    Delay & $8.2\%$  & $13.4\%$ & $19/43$ & $39/43$ \\
    Power & $31.3\%$ & $35.8\%$ &  $4/43$ & $13/43$ \\
    \bottomrule
  \end{tabular}
\end{table}

The full per-design CSV (\texttt{reference\_ppa\_measured.csv}) is
released alongside the codebase.

\section{KL-Budget Ablation Trajectories}
\label{app:kl_budget_trajectories}

\begin{figure}[h]
  \centering
  \includegraphics[width=\linewidth]{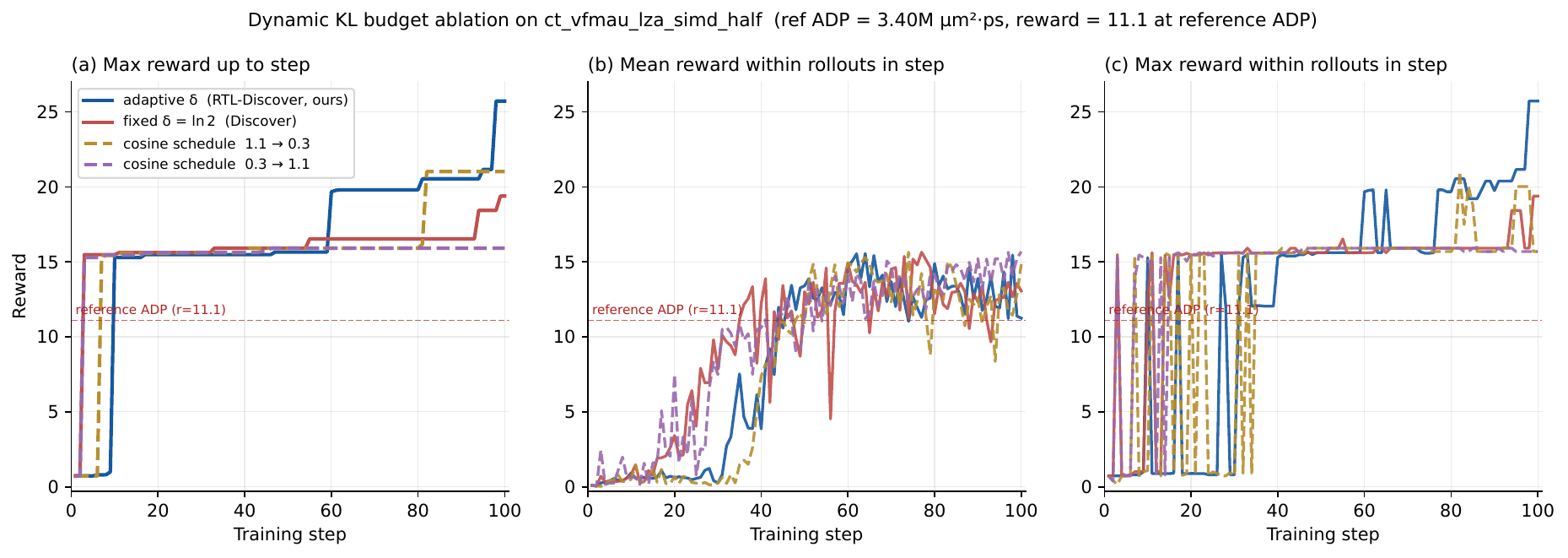}
  \caption{%
    KL-budget ablation trajectories on
    \texttt{ct\_vfmau\_lza\_simd\_half}, in the same three-panel
    layout as \cref{fig:ablations}.
    Adaptive $\delta$ (dark blue, RTL-Discover) keeps a strictly
    higher upper-tail reward than all three alternatives throughout
    training (panel~c) and is the only strategy whose best-so-far
    reward (panel~a) crosses $r=20$.
    The cosine $1.1\!\to\!0.3$ schedule recovers a substantial
    fraction of the gap, supporting the interpretation that
    early-large / late-small $\delta$ is a useful inductive bias for
    RTL search; the reverse schedule is uniformly worse.
    The fixed Discover-style $\delta = \ln 2$ falls in between but
    plateaus before reaching the reference-ADP line on (a).
    Headline numbers are in \cref{tab:ablations}
    (\cref{sec:c910_kl_budget}).
  }
  \label{fig:kl_budget_ablations}
\end{figure}

\section{Multi-Seed and Multi-Task Replication of the Adaptive KL-Budget Controller}
\label{app:kl_budget_multiseed}

\Cref{sec:c910_kl_budget} reports the adaptive vs.\ fixed $\delta = \ln 2$
contrast at a single seed on \texttt{ct\_vfmau\_lza\_simd\_half}.  We
extend that contrast in two directions: (i) a four-seed paired
replication on the same problem, and (ii) a single-seed case study on a
second C910 unit, \texttt{ct\_vfdsu\_fadd\_close\_s0\_d}.  Both arms
share the recipe of \cref{sec:c910_kl_budget} (PUCT + entropic, sky130,
KL penalty active, $100$ RL steps, $4$ parents $\times\,8$ rollouts) and
differ only in the algorithmic switch
\texttt{algorithm.ttt\_entropic\_kl\_budget\_mode}.  Raw extracts,
statistics scripts, and figures are released under
\texttt{experiment/checklist/multi-seed-c910/}.

\begin{figure}[h]
  \centering
  \includegraphics[width=0.49\linewidth]{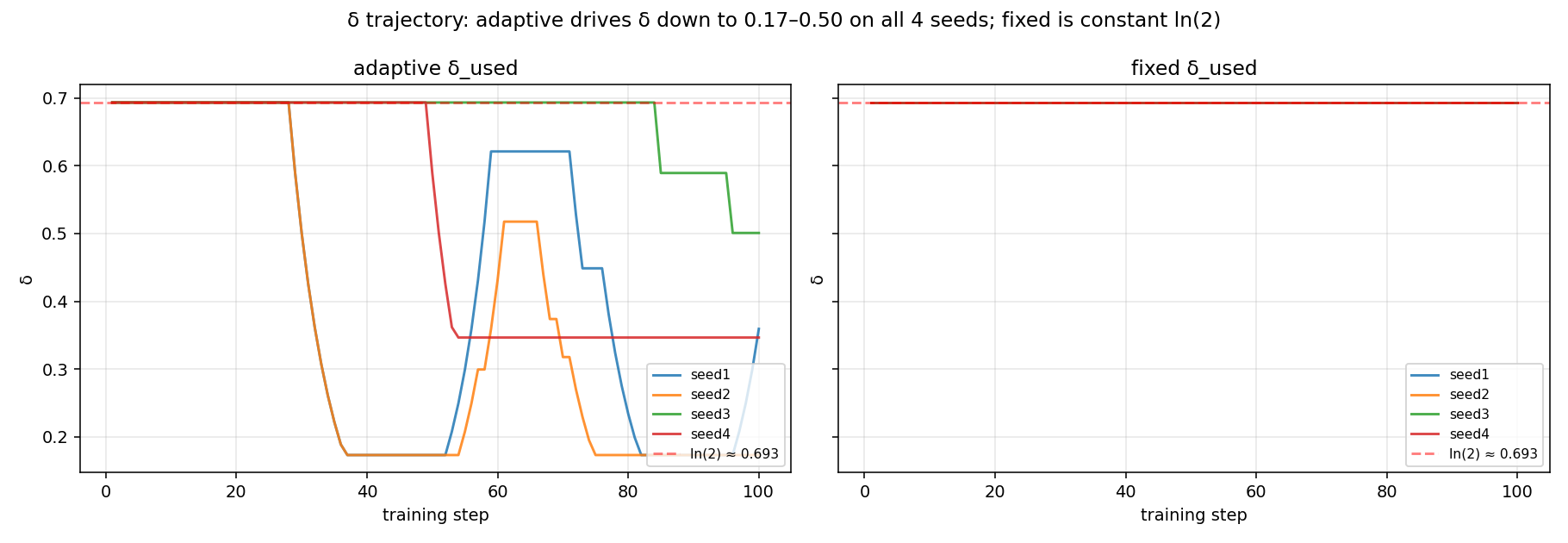}\hfill
  \includegraphics[width=0.49\linewidth]{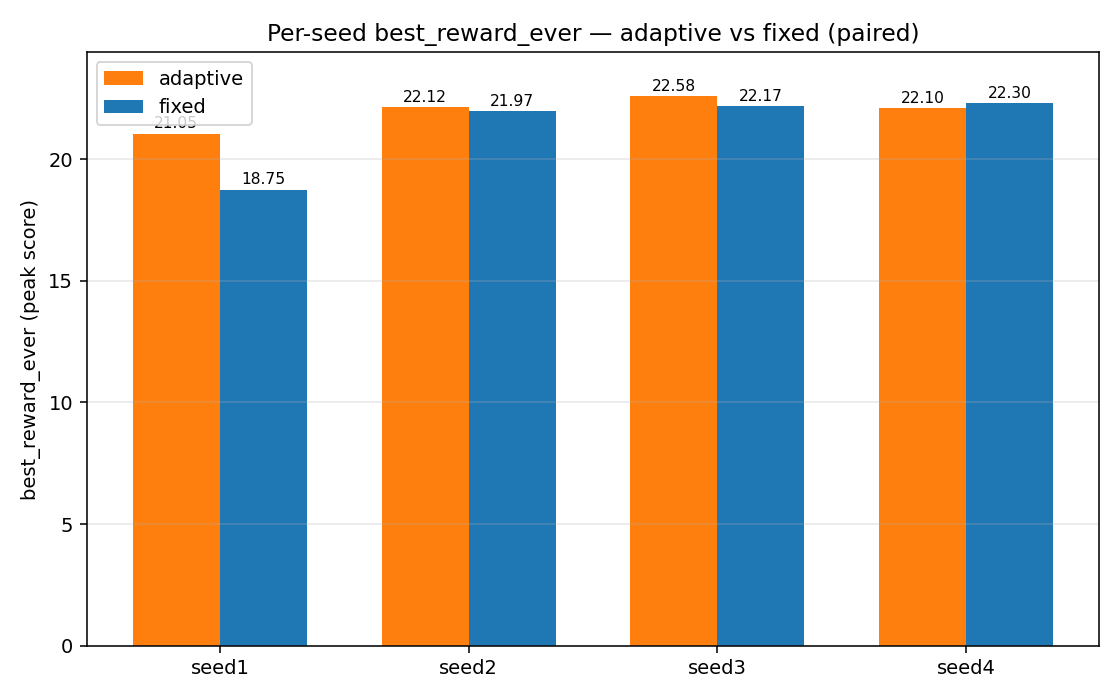}
  \caption{%
    LZA \texttt{simd\_half}, $n = 4$ paired.
    \textbf{Left:} per-step $\delta_t$ for all four adaptive seeds and
    the fixed $\delta = \ln 2$ baseline.  Adaptive consistently drives
    $\delta$ below $\ln 2$ across seeds.
    \textbf{Right:} per-seed \texttt{best\_reward\_ever} bars
    (paired).  Adaptive wins $3 / 4$ but the gap is within seed-wise
    noise.
  }
  \label{fig:multiseed_lza}
\end{figure}

\paragraph{LZA \texttt{simd\_half} (n = 4 paired).}
\Cref{tab:multiseed_lza} shows the four headline metrics paired across
seeds.  No metric reaches $p < 0.05$ under either paired $t$-test or
Wilcoxon at $n = 4$; the unpaired Welch test on
\texttt{best\_reward\_ever} returns $p = 0.51$.  The mean advantage of
$+0.67$ on \texttt{best\_reward\_ever} is robust to dropping any single
seed (leave-one-out $\Delta \in [+0.12, +0.95]$, all positive), but the
gap remains within seed-wise noise.  Adaptive does, however, reduce
seed-wise standard deviation by $\sim\!2.6\times$
($0.56$ vs.\ $1.48$) and reaches score $\geq 20$ on $4 / 4$ seeds while
fixed reaches it on $3 / 4$.  Across all four adaptive seeds the
controller drives $\delta$ \emph{below} $\ln 2$, with per-seed minima in
$[0.17, 0.50]$ (\cref{fig:multiseed_lza}, top).

\begin{table}[h]
  \centering
  \caption{%
    Paired multi-seed comparison on \texttt{ct\_vfmau\_lza\_simd\_half}
    (sky130, $n = 4$ adaptive vs.\ $4$ fixed, $100$ RL steps each).
    Mean $\pm$ std across seeds; $p$-values are paired $t$-test.
    No metric reaches $p < 0.05$.
  }
  \label{tab:multiseed_lza}
  \small
  \begin{tabular}{lccccc}
    \toprule
    Metric & Direction & adaptive & fixed $\delta = \ln 2$ & $\Delta$ & paired $p$ \\
    \midrule
    \texttt{best\_reward\_ever}        & $\uparrow$ & $\mathbf{21.96 \pm 0.56}$ & $21.30 \pm 1.48$ & $+0.67$  & $0.32$ \\
    \texttt{end\_reward\_last20}       & $\uparrow$ & $19.01 \pm 1.60$ & $\mathbf{19.44 \pm 2.00}$ & $-0.43$  & $0.57$ \\
    \texttt{step\_to\_score} $\geq 20$ & $\downarrow$ & $\mathbf{44.5 \pm 33.3}$ & $69.5 \pm 77.7$ & $-25.0$ & $0.42$ \\
    \texttt{const\_group\_frac\_last20} & $\downarrow$ & $\mathbf{0.46 \pm 0.18}$ & $0.52 \pm 0.09$ & $-0.06$ & $0.58$ \\
    \bottomrule
  \end{tabular}
\end{table}

The four adaptive runs were not originally generated as seeds $1$--$4$;
they are historical runs treated as a four-seed pool for this
analysis.  An exhaustive enumeration of all $24$ adaptive-to-fixed
permutations yields paired-$t$ $p \in [0.31, 0.59]$ with $0/24$
significant at $p < 0.10$, so the headline conclusion is invariant to
pairing choice.

\begin{figure}[h]
  \centering
  \includegraphics[width=0.49\linewidth]{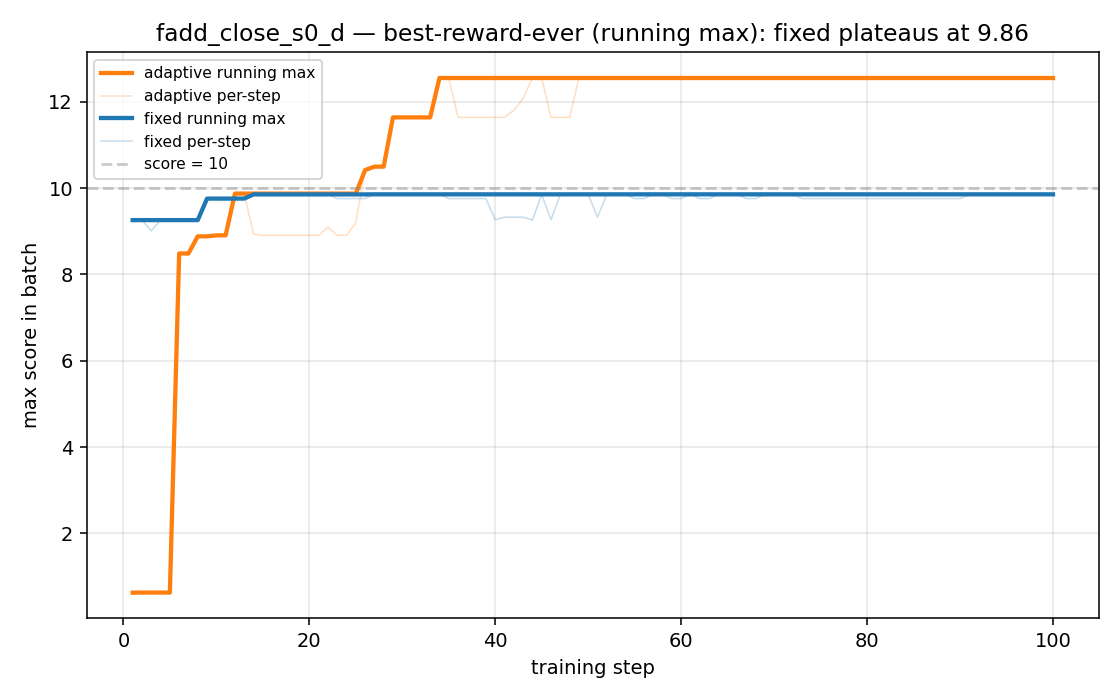}\hfill
  \includegraphics[width=0.49\linewidth]{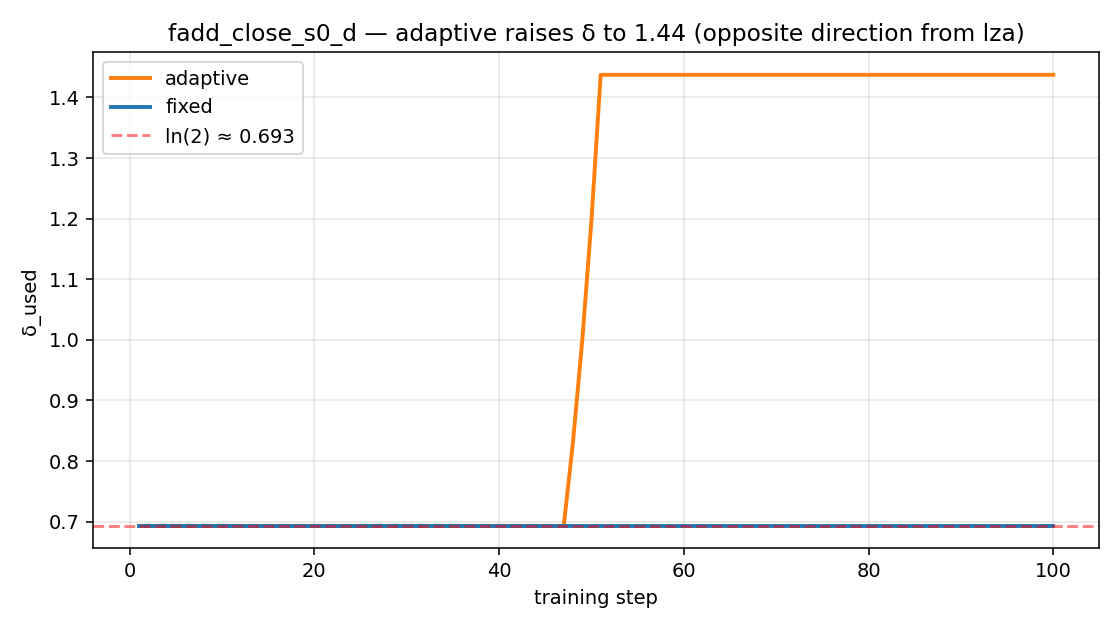}
  \caption{%
    \texttt{fadd\_close\_s0\_d}, $n = 1$ vs.\ $1$.
    \textbf{Left:} running-max best-in-batch reward.  Fixed plateaus at
    $9.86$; adaptive escapes around step $30$ and stabilises near
    $12.5$.
    \textbf{Right:} $\delta_t$ trajectory.  Fixed stays at
    $\ln 2 \approx 0.693$; adaptive raises $\delta$ to $1.44$ at step
    $\sim\!50$, \emph{after} the policy breakthrough.
  }
  \label{fig:multiseed_fadd}
\end{figure}

\paragraph{\texttt{fadd\_close\_s0\_d} ($n = 1$).}
A single-seed case study on a second C910 unit shows a qualitatively
different picture: the fixed arm plateaus at \texttt{best\_reward} =
$9.86$ from step $\sim\!20$ onwards and never crosses $10$, while the
adaptive arm reaches $12.55$ (peak) and $12.33$ (end-of-run mean over
the last $20$ steps), a $+27\%$ / $+34\%$ relative gap.  Crucially,
the controller \emph{raises} $\delta$ above $\ln 2$ on this problem
(to $1.44$, see \cref{fig:multiseed_fadd}, right) -- the opposite
direction from LZA -- demonstrating that the four EMA signals of
\cref{sec:advantage} respond to task-specific structure rather than
moving in a fixed direction.

\paragraph{What this experiment supports.}
(1) The adaptive controller is task-responsive in direction -- it
lowers $\delta$ on LZA and raises it on fadd.
(2) Adaptive matches or exceeds fixed on every aggregate metric we
report ($+0.67$ peak and $-25$ steps to score $\geq 20$ on LZA;
$+2.69$ peak and $+3.11$ end-of-run on fadd).
(3) Seed-wise variance on LZA is reduced $\sim\!2.6\times$ and the
worst-case fixed seed (peak $18.75$, fails to cross score $20$) has no
adaptive counterpart.

\paragraph{What this experiment does \emph{not} support.}
(1) Statistical significance: no metric reaches $p < 0.05$ on LZA at
$n = 4$, and fadd is $n = 1$ and admits no test.
(2) Causal attribution on fadd: the $\delta$ raise occurs at step
$\sim\!50$, after the policy has already plateaued at the new peak
near step $30$.  In the first $50$ steps both arms run with $\delta
= \ln 2$ and the divergence is from PUCT / vLLM stochasticity, not the
scheduler.
(3) A peak-performance claim: the LZA $+0.67$ falls inside the
$[15.6, 25.7]$ development-time spread reported in
\cref{sec:conclusion}, and fadd is a single trial.

We accordingly frame the adaptive controller in the body of the paper
as a robustness and task-responsiveness observation rather than a
peak-performance win.  Re-running fadd at $n \geq 4$ and extending LZA
to $n \geq 8$ remain the obvious follow-ups.

\section{C910 LZA Case Study: Baseline, Optimized Code, and Testbench}
\label{app:c910_case}

This appendix documents the concrete artifact behind the C910 numbers in
\cref{sec:c910_kl_budget,sec:c910_kl_budget,app:kl_budget_multiseed}: the
baseline RTL we start from, the best discovered candidate
(\texttt{best\_reward\_ever} = $25.72$, $A{\cdot}D = 1.38{\times}10^{6}$
vs.\ baseline $3.40{\times}10^{6}$, i.e.\ $-59.4\%$ ADP at sky130 HD), and
the verification harness that gates every rollout.

\subsection{Baseline: Xuantie OpenC910 \texttt{ct\_vfmau\_lza\_simd\_half}}

The baseline is the upstream RTL from the open-sourced Xuantie C910 core
(\texttt{ct\_vfmau\_lza\_simd\_half.v}), which implements a $24$-bit
leading-zero anticipator for the SIMD half-precision lane of the VFMAU.
The module has two stages:

\paragraph{Stage~1: pre-encode.} Carry signals
$P{=}\mathrm{summand}\oplus\mathrm{addend}$, $G{=}\mathrm{summand}\,\&\,\mathrm{addend}$,
$D{=}\overline{\mathrm{summand}\,|\,\mathrm{addend}}$ are formed bitwise,
followed by a vectorized $24$-bit pre-decode \texttt{lza\_precod[23:0]}
with three boundary cases (LSB, MSB, and the bulk $[22{:}1]$ slice
gated on \texttt{sub\_vld}).  Both arms keep this stage byte-for-byte
identical; the optimization budget is entirely in stage~2.

\paragraph{Stage~2 (baseline): flat $24$-way \texttt{casez} priority
encoder.}  The leading-one position is decoded by a single
\texttt{always @(*) casez} block with $24$ one-hot patterns of the form
\texttt{24'b1???\dots} $\to$ \texttt{5'd0}, \texttt{24'b01??\dots} $\to$
\texttt{5'd1}, $\dots$, \texttt{24'b\dots001} $\to$ \texttt{5'd23}, plus
a default $\texttt{5'd24}$.  Yosys synthesizes this into a $24$-input
priority chain whose critical path traverses all $24$ \texttt{casez}
arms before the $5$-bit encoder; together with the (already large)
pre-encode this is what gives the baseline ADP of $3.40{\times}10^{6}$.

\subsection{Best discovered candidate}

The best rollout under our adaptive-$\delta$ recipe keeps stage~1
unchanged and rewrites stage~2 as a \emph{hierarchical $4$-bit grouped}
priority encoder (\cref{lst:c910_best}).  Three changes drive the ADP
reduction:
\begin{enumerate}
  \item \textbf{Group-then-encode.}  \texttt{lza\_precod[23:0]} is sliced
  into six $4$-bit groups
  ($\{23{:}20\},\{19{:}16\},\dots,\{3{:}0\}$).
  Within a group the $4$-way priority over a $4$-bit one-hot vector
  collapses to a depth-$3$ ternary cascade
  \texttt{a[3]?\,k:\,a[2]?\,k{+}1:\,a[1]?\,k{+}2:\,a[0]?\,k{+}3:\,5'd24},
  which Yosys maps to four $2$-input gates per group instead of the
  shared $24$-priority chain.

  \item \textbf{Group-validity reuse.}  Six instances of the existing
  C910 sub-cell \texttt{ct\_vfmau\_lza\_42} (a 4:2 LZA compressor
  already shipped in the \texttt{deps/} directory) compute a
  \texttt{lza\_vld} flag per group; the outer cascade picks the
  highest-index valid group, so each $4$-bit decoder fires only when
  its group is selected.  The \texttt{lza\_p0}/\texttt{lza\_p1} outputs
  of the compressor are deliberately left unconnected --- the policy
  reuses the cell only for its valid bit.  This converts the
  baseline's monolithic priority chain into a $\log_4(24)$-style
  two-level structure, which is the dominant source of the delay
  reduction ($2191\to 1315\,\mathrm{ps}$, $-40\%$).

  \item \textbf{Combinational, not registered.}  The baseline declares
  \texttt{reg [4:0] lza\_result} and drives it from an
  \texttt{always @(*)} block; the candidate uses a single
  \texttt{assign} with a nested ternary, which is functionally
  equivalent (no clock anywhere in the spec) but lets Yosys schedule
  the decoder under one cone of logic with the group-validity lookups,
  trimming both area ($1553\to 1051\,\mu\mathrm{m}^{2}$, $-32\%$) and
  the fan-in to the final mux.
\end{enumerate}

The interface (port list, widths, directions) is preserved exactly, so
the candidate is a drop-in replacement.  This module-internal
restructuring is the only change in stage~2.

\begin{lstlisting}[language=Verilog,
                   caption={Stage~2 of the best candidate
                            (\texttt{ttt\_c910\_state\_pool/latest.json},
                            state \#304, $A{\cdot}D = 1.38{\times}10^{6}$).
                            Stage~1 (\texttt{carry\_p/g/d}, \texttt{lza\_precod})
                            is unchanged from the baseline and elided.},
                   label=lst:c910_best,
                   basicstyle=\scriptsize\ttfamily,
                   breaklines=true,
                   columns=fullflexible]
// 4-bit grouping of lza_precod
wire [3:0] lza_4_23_20 = lza_precod[23:20];
wire [3:0] lza_4_19_16 = lza_precod[19:16];
wire [3:0] lza_4_15_12 = lza_precod[15:12];
wire [3:0] lza_4_11_8  = lza_precod[11:8];
wire [3:0] lza_4_7_4   = lza_precod[7:4];
wire [3:0] lza_4_3_0   = lza_precod[3:0];

// Reuse C910 LZA 4:2 compressor for the per-group valid flag only
ct_vfmau_lza_42 lza_42_23_20 (.lza_precod(lza_4_23_20),
                              .lza_p0(), .lza_p1(), .lza_vld());
// ... five more identical instances for the remaining groups ...

// Hierarchical priority: highest-index valid group wins
assign lza_result =
   (lza_42_23_20.lza_vld) ?
       (lza_4_23_20[3] ? 5'd0  : lza_4_23_20[2] ? 5'd1  :
        lza_4_23_20[1] ? 5'd2  : lza_4_23_20[0] ? 5'd3  : 5'd24) :
   (lza_42_19_16.lza_vld) ?
       (lza_4_19_16[3] ? 5'd4  : lza_4_19_16[2] ? 5'd5  :
        lza_4_19_16[1] ? 5'd6  : lza_4_19_16[0] ? 5'd7  : 5'd24) :
   /* ... four more group arms ... */
   (lza_42_3_0.lza_vld) ?
       (lza_4_3_0[3]   ? 5'd20 : lza_4_3_0[2]   ? 5'd21 :
        lza_4_3_0[1]   ? 5'd22 : lza_4_3_0[0]   ? 5'd23 : 5'd24) :
   5'd24;

assign lza_result_zero = ~|lza_precod[23:0];
\end{lstlisting}

\subsection{Verification harness}
\label{app:c910_testbench}

Every rollout is gated by an \texttt{iverilog} testbench that
co-instantiates the candidate and a renamed copy of the baseline
(\texttt{ct\_vfmau\_lza\_simd\_half\_ref}) and compares
\texttt{lza\_result} and \texttt{lza\_result\_zero} cycle by cycle.
The harness contains $5$ phases ($\sim\!1{,}050$ vectors total):
\begin{itemize}
  \item \textbf{Edge cases (sub\_vld\,=\,0, addition mode).}  Both
  operands zero, both all-ones, MSB-only and LSB-only patterns,
  one operand zero with the other all-ones, alternating
  $0xAAAAAA{/}0x555555$, walking-1 over all $24$ bit positions on each
  operand, adjacent-bit carry-propagate patterns, and the three
  saturated carry chains $P{=}\mathbf{1},G{=}\mathbf{1},D{=}\mathbf{1}$.
  \item \textbf{Edge cases (sub\_vld\,=\,1, subtraction mode).}  The
  same zero / all-ones / boundary / walking-1 / alternating set,
  exercising the \texttt{sub\_vld}-dependent branches of
  \texttt{lza\_precod[0]} and \texttt{lza\_precod[23]}.
  \item \textbf{Random, $\mathbf{800}$ vectors.}  $400$ with
  \texttt{sub\_vld\,=\,0} and $400$ with \texttt{sub\_vld\,=\,1}, drawn
  from \texttt{\$random(seed)} with seed $42$.
  \item \textbf{Sparse / mixed.}  $100$ vectors with
  \texttt{sub\_vld} sampled per trial and only $2$--$3$ random bits
  set across the two operands, to stress the priority encoder under
  near-degenerate inputs.
  \item \textbf{Close-path.}  $100$ vectors with
  \texttt{addend} = \texttt{summand} XOR-perturbed at one random
  bit position under \texttt{sub\_vld\,=\,1}; this is the regime
  where leading-zero anticipation dominates the floating-point
  add-round critical path and is the original engineering motivation
  for the LZA cell.
\end{itemize}
A run is accepted only when all $\sim\!1{,}050$ comparisons pass
($\texttt{fail\_count}=0$, ``Your Design Passed''); any single
mismatch zeroes the functional reward and the candidate is rejected
before PPA synthesis is even invoked.  This is what makes the
$-59.4\%$ ADP claim a functional-equivalence claim against the
upstream Xuantie cell rather than a synthesis-only PPA claim.

\section{Hyperparameter Summary}
\label{app:hyperparams}

\begin{table}[h]
  \centering
  \caption{TTT-RTL hyperparameter configuration used in all experiments.}
  \label{tab:hyperparams}
  \begin{tabularx}{\linewidth}{lX}
    \toprule
    Hyperparameter & Value \\
    \midrule
    Base model & RTLLM v2.0: Qwen3-8B + \texttt{ttt-rtl-sft} step 18 (\cref{app:sft}); C910 LZA: raw Qwen3-8B (no SFT) \\
    Gradient steps per problem & 100 \\
    Batch size (parent states) & 4 \\
    Rollouts per prompt ($n$) & 4 (RTLLM v2.0 main runs) / 8 (C910 LZA ablations) \\
    PUCT exploration coefficient ($c$) & 1.0 \\
    State pool size ($C_{\max}$) & 500 \\
    Top-$k$ children per parent & 2 \\
    Reward weight $\omega_{\text{syn}}$ & 0.1 \\
    Reward weight $\omega_{\text{func}}$ & 1.0 \\
    Reward weight $\omega_{\text{ppa}}$ & 10.0 \\
    KL budget $\delta_t$ for $\betastar$ search & RTLLM v2.0: fixed $\delta = \ln 2$ (TTT-Discover canonical); C910 LZA: per-row, see \cref{tab:ablations} (incl.\ adaptive controller of \cref{tab:adaptive_hparams}, range $[0.25\ln 2,4\ln 2]$) \\
    Technology library & Nangate 45\,nm (RTLLM v2.0) / Sky130 HD (C910 LZA) \\
    Synthesis tool & Yosys \\
    Functional simulation & iverilog / vvp \\
    Training framework & \texttt{verl} \citep{sheng2024hybridflow} \\
    \bottomrule
  \end{tabularx}
\end{table}

\section{SFT Warm-up Recipe}
\label{app:sft}

The \texttt{ttt-rtl-sft} checkpoint that initializes the policy model is
a lightweight format-and-style warm-up applied to Qwen3-8B \emph{before}
any test-time RL.  Its sole purpose is to make the base model reliably
emit Verilog inside the expected \texttt{<think>...</think>} block and
\texttt{```verilog} fences while respecting the prompt structure of
\cref{app:prompts}; it is \textbf{not} intended as a knowledge
distillation step that would short-circuit the role of test-time
training.

\paragraph{Data source.}
The SFT corpus is $5{,}000$ uniformly sampled rows from the public
CoDeV-R1 distillation dataset \citep{codev_r1_dataset} (87K Verilog
reasoning-chain records in total), with the original
\texttt{<think>}/\texttt{<answer>} tags rewritten to match our
\texttt{<think>} convention.  No RTLLM~v2.0 problems are used at SFT
time --- the warm-up is purely an out-of-distribution format-and-style
prior, and all benchmark exposure happens during test-time RL.  The
corpus is in the verl multi-turn \texttt{messages} format with a
$95/5$ train/val split.

\paragraph{Training.}
We train Qwen3-8B for $3$ epochs at learning rate $10^{-5}$ with FSDP,
global batch size $64$, max sequence length $16{,}384$, on $8\times$
A800 GPUs.  All RL experiments in this paper start from the
global-step-$18$ checkpoint, which corresponds to roughly $1{,}152$
examples seen ($\sim\!0.23$ epochs over the corpus): on average each
training row has been seen \emph{at most once}.  This minimal warm-up
is intentional --- in pilot runs longer SFT degraded downstream
exploration during RL --- and the goal is only to lock in
response format, not to teach problem-specific solutions.

\paragraph{Is SFT or test-time RL doing the work?}
Two factors bound the risk that SFT, rather than test-time RL, drives
the headline numbers.
(1) The $\sim\!0.23$-epoch budget on a corpus that contains no RTLLM
v2.0 problems is too small for the model to memorize benchmark
solutions: the warm-up shifts response \emph{format}, not benchmark
answers.
(2) The C910 LZA experiments do not use the SFT checkpoint at all:
they initialize from the raw Qwen3-8B base model
(\cref{tab:hyperparams}).  The \emph{Best-of-$N$} row of
\cref{tab:ablations} freezes that raw base policy and draws $N=3200$
samples on \texttt{ct\_vfmau\_lza\_simd\_half}; it never produces a
single functionally correct design within the budget.  Since SFT plays
no role on C910, the $-59.4\%$ ADP improvement on this unit is
attributable to test-time RL on top of the off-the-shelf backbone.
The released artifact ships the SFT
configuration and merge script so reviewers can re-run the warm-up
under any policy of choice.

\section{Prompt Templates}
\label{app:prompts}

\paragraph{System prompt.}
The system prompt is fixed for all problems:
\begin{quote}
\small
\texttt{You are an expert Verilog RTL designer.}\\
\texttt{When reasoning, use <think>...</think> tags.}\\
\texttt{Output only Verilog code within ```verilog fences.}
\end{quote}

The templates below are the RTLLM~v2.0 main-run prompts in which the
PPA product $M = A \cdot D \cdot P$ is shown to the model.  In the
C910 LZA ablation (where the OpenSTA power column is not collected)
the same templates degenerate to ADP ($M = A \cdot D$); we use the
generic placeholder ``\texttt{\{M\}}'' below to make this dual usage
explicit.

\paragraph{Root state user prompt.}
\begin{quote}
\small
\texttt{\#\# Reference Implementation}\\
\texttt{Below is a known working implementation that synthesizes to}\\
\texttt{area=\{A\}$\mu$m\^{}2, delay=\{D\}ps, power=\{P\}$\mu$W (PPA-product=\{M\}).}\\
\texttt{Your task is to produce a functionally correct implementation}\\
\texttt{with lower PPA-product (Area $\times$ Delay $\times$ Power).}\\
\texttt{```verilog}\\
\texttt{\{reference\_code\}}\\
\texttt{```}\\[2pt]
\texttt{\#\# Design Specification}\\
\texttt{\{specification\}}\\[2pt]
\texttt{Write a complete Verilog module with PPA-product lower than \{M\}.}\\
\texttt{Output only Verilog code within ```verilog fences.}
\end{quote}

\paragraph{Non-root state user prompt.}
The non-root prompt appends the following block after the reference and specification:
\begin{quote}
\small
\texttt{\#\# Previous Attempt}\\
\texttt{```verilog}\\
\texttt{\{parent\_code\}}\\
\texttt{```}\\[2pt]
\texttt{\#\# Feedback from Previous Attempt}\\
\texttt{- Syntax: PASS}\\
\texttt{- Functional Test: PASS}\\
\texttt{- Synthesis: area=\{A\}$\mu$m\^{}2, delay=\{D\}ps, power=\{P\}$\mu$W, PPA-product=\{M\}}\\
\texttt{  Previous PPA-product: \{prev\_M\} -> current: \{M\} (improved by \{$\Delta$\})}\\[2pt]
\texttt{You are iteratively optimizing PPA-product (lower is better).}\\
\texttt{Produce a design with PPA-product lower than \{M\}.}
\end{quote}

\section{Entropic Adaptive Beta: Implementation Detail}
\label{app:beta}

\paragraph{Stage-1 syntax score.}
On \texttt{iverilog} failure the syntax reward is
$r_{\text{syn}} = 1 / (1 + n_{\text{err}})$, additionally multiplied
by $0.3$ when the log contains a port-binding keyword (\texttt{port},
\texttt{unknown module}, \texttt{not a module}), and falling back to
$0.5$ when the log cannot be parsed.

\paragraph{Binary search for $\betastar$.}
The binary search for $\betastar$ operates on $[10^{-6}, 10^{6}]$ with 64 iterations
at the current step's KL budget $\delta_t$.
TTT-Discover \citep{yuksekgonul2026discover} sets $\delta_t \equiv \ln 2$ as a
constant; we report this as the \emph{fixed} baseline in the KL-budget ablation
(\cref{tab:ablations}).
RTL-Discover replaces the constant with the adaptive controller of
\cref{sec:advantage}; its hyperparameters are listed in
\cref{tab:adaptive_hparams}.
Intuitively, the constant $\ln 2$ enforces that the group's softmax
distribution is no more concentrated than a Bernoulli(0.5) distribution,
preventing the policy gradient from collapsing to a single rollout even
when reward differences are large; the adaptive controller relaxes this
hard cap and instead modulates $\delta$ between
$0.25\ln 2$ and $4\ln 2$ in response to the four EMA signals described
in \cref{sec:advantage}.

\paragraph{Signals and EMA update.}
At step $t$, given the batch of $G$ groups produced by $\delta_{t-1}$,
the controller measures four scalar signals: (i) the average
policy-vs-reference KL $\mathrm{KL}_{\mathrm{ref}}$, (ii) the
effective number of distinct rollouts averaged across groups,
$\mathrm{eff\text{-}n} = \frac{1}{G}\sum_{g=1}^{G}(\sum_i q_{g,i}^2)^{-1}$,
(iii) the fraction of groups with degenerate (constant) reward,
$\mathrm{const\text{-}frac}$, and (iv) the binary-search saturation
rate $\beta_{\max}\text{-}\mathrm{rate}$ (the fraction of groups whose
$\betastar$ hits the search bound $10^{6}$).
Each signal is smoothed by a single EMA with mixing $\alpha = 0.30$:
\begin{equation}
  \bar{x}_t = (1 - \alpha)\,\bar{x}_{t-1} + \alpha\, x_t,
  \qquad x \in \{\mathrm{KL}_{\mathrm{ref}},\,\mathrm{eff\text{-}n},\,
                  \mathrm{const\text{-}frac},\,\beta_{\max}\text{-}\mathrm{rate}\}.
  \label{eq:adaptive-ema}
\end{equation}
A separate slower EMA with $\alpha_r = 0.10$ tracks the running peak
reward $\bar{r}_t$ used by the stagnation counter
$N_{\mathrm{noimp}}$, which is incremented when the batch maximum
fails to exceed $\bar{r}_{t-1}$ by at least $\Delta r_{\min} = 0.02$
and reset to $0$ whenever a new peak is observed.

\paragraph{Priority ladder.}
For the first $T_{\mathrm{warm}} = 20$ steps only the KL brake (P1) is
active, with $\delta_t \equiv \delta_0$ otherwise.  Past warm-up, the
ladder evaluates P1--P4 in order and applies the assignment of the
\emph{first} rule whose guard holds; if none holds, $\delta_t \leftarrow
\delta_{t-1}$ (\textbf{hold}).
\begin{align}
  \text{(P1) KL brake:}\quad &
    \overline{\mathrm{KL}_{\mathrm{ref}}}_t > \tau_{\mathrm{KL}}
    \;\Rightarrow\; \delta_t \leftarrow \delta_0,
    \label{eq:p1}\\
  \text{(P2) winner-take-all:}\quad &
    \overline{\beta_{\max}\text{-}\mathrm{rate}}_t > \tau_{\beta}
    \;\wedge\;
    \overline{\mathrm{eff\text{-}n}}_t < \tau_n^{\mathrm{wta}}
    \;\Rightarrow\; \delta_t \leftarrow \max(\rho_{-}\,\delta_{t-1},\,\delta_{\min}),
    \label{eq:p2}\\
  \text{(P3) stagnation:}\quad &
    N_{\mathrm{noimp}} \ge T_{\mathrm{stag}}
    \;\wedge\;
    \overline{\mathrm{const\text{-}frac}}_t < \tau_{\mathrm{const}}
    \;\wedge\;
    \overline{\mathrm{eff\text{-}n}}_t < \tau_n^{\mathrm{stag}}
    \;\Rightarrow\; \delta_t \leftarrow \max(\rho_{-}\,\delta_{t-1},\,\delta_{\min}),
    \label{eq:p3}\\
  \text{(P4) over-exploring:}\quad &
    \overline{\mathrm{eff\text{-}n}}_t > \tau_n
    \;\wedge\;
    \overline{\mathrm{const\text{-}frac}}_t < \tau_{\mathrm{const}}^{\mathrm{lo}}
    \;\wedge\;
    N_{\mathrm{noimp}} < T_{\mathrm{plat}}
    \;\Rightarrow\; \delta_t \leftarrow \min(\rho_{+}\,\delta_{t-1},\,\delta_{\max}).
    \label{eq:p4}
\end{align}
The \emph{eff-n} joint gates on P2 and P3 ($\tau_n^{\mathrm{wta}}$,
$\tau_n^{\mathrm{stag}}$) are essential: they prevent the controller
from shrinking $\delta$ when group weights are already near uniform
(a flat-reward regime that would collapse the advantages to zero on
further shrinkage), and instead require that the rule that fires
acts on a genuinely concentrated group.

\begin{table}[h]
  \centering
  \caption{%
    Hyperparameters of the adaptive KL-budget controller
    (\cref{sec:advantage}).
    All values are held fixed across all problems and PDKs reported in this
    paper.
    Symbols match the priority-ladder description in
    \cref{sec:advantage}.
  }
  \label{tab:adaptive_hparams}
  \small
  \begin{tabular}{llr}
    \toprule
    Symbol & Description & Value \\
    \midrule
    $\delta_0$ & Init / KL-brake reset value & $\ln 2 \approx 0.693$ \\
    $\delta_{\min}$ & Lower bound of $\delta_t$ & $0.25\ln 2 \approx 0.173$ \\
    $\delta_{\max}$ & Upper bound of $\delta_t$ & $4\ln 2 \approx 2.773$ \\
    $\rho_{-}$ & Shrink factor (P2, P3) & $0.85$ \\
    $\rho_{+}$ & Grow factor (P4) & $1.20$ \\
    $\alpha$ & EMA mixing for KL / eff-$n$ / const-frac / $\beta$-rate & $0.30$ \\
    $\alpha_{r}$ & EMA mixing for reward peak tracking & $0.10$ \\
    $\tau_{\mathrm{KL}}$ & P1 KL-brake threshold & $0.40$ \\
    $\tau_{\beta}$ & P2 winner-take-all $\beta$-rate threshold & $0.60$ \\
    $\tau_n^{\mathrm{wta}}$ & P2 eff-$n$ joint gate (sharp groups only) & $3.5$ \\
    $T_{\mathrm{stag}}$ & P3 stagnation steps & $15$ \\
    $\Delta r_{\min}$ & P3 minimum relative improvement & $0.02$ \\
    $\tau_{\mathrm{const}}$ & P3 const-frac guard (signal exists) & $0.70$ \\
    $\tau_n^{\mathrm{stag}}$ & P3 eff-$n$ joint gate (sharp groups only) & $4.0$ \\
    $\tau_n$ & P4 effective-$n$ threshold & $5.5$ \\
    $\tau_{\mathrm{const}}^{\mathrm{lo}}$ & P4 const-frac threshold & $0.30$ \\
    $T_{\mathrm{plat}}$ & P4 plateau guard (steps) & $5$ \\
    $T_{\mathrm{warm}}$ & Warm-up steps (P1 only) & $20$ \\
    \bottomrule
  \end{tabular}
\end{table}

\clearpage
\section*{NeurIPS Paper Checklist}

\begin{enumerate}

\item {\bf Claims}\\
Question: Do the main claims made in the abstract and introduction accurately reflect the paper's contributions and scope?\\
Answer: \answerYes{}\\
Justification: The abstract and introduction state three contributions: (i) a per-design test-time training framework that closes the LLM--EDA loop (\cref{sec:method}), (ii) external comparison on RTLLM~v2.0 reporting a $65.1\%$ geomean PPA-product reduction vs.\ the v2.0 reference, against $26.1\%$ for the strongest published baseline (\cref{tab:main_results}, \cref{sec:main_results}), and (iii) an industrial C910 LZA case study with a $59.4\%$ ADP reduction and component ablations (\cref{tab:ablations}, \cref{sec:ablations}). The adaptive KL-budget controller is framed as a robustness/task-responsiveness observation rather than a peak-performance win, matching the multi-seed evidence in \cref{app:kl_budget_multiseed}.

\item {\bf Limitations}\\
Question: Does the paper discuss the limitations of the work performed by the authors?\\
Answer: \answerYes{}\\
Justification: \textbf{(L1) Single-seed main results.} RTLLM~v2.0 and the C910 main-table results are single-seed (matching the published agent baselines, none of which report seed variance); a four-seed paired replication on LZA \texttt{simd\_half} and a single-seed case study on \texttt{ct\_vfdsu\_fadd\_close\_s0\_d} (\cref{app:kl_budget_multiseed}) confirm direction and a ${\sim}2.6\times$ seed-wise variance reduction but do not reach $p<0.05$ at $n=4$, so per-row gaps in \cref{tab:ablations} should be read as ranking evidence rather than tight effect sizes. \textbf{(L2) Simulation-only correctness.} Designs are validated against the RTLLM testbench, not formal equivalence; a Yosys \texttt{eqy} check is the natural next step, as is timing-accurate synthesis (OpenROAD~\citep{ajayi2019openroad} or commercial flows) for more reliable PPA estimates. \textbf{(L3) Internal ablation of the controller.} We do not yet report a leave-one-rule-out study of P1--P4 within the adaptive controller, so we cannot claim which rule drives the gain. \cref{app:kl_budget_multiseed} additionally documents the controller's negative results: no metric reaches $p<0.05$ at $n=4$ on LZA, the fadd $\delta$ raise post-dates the breakthrough, and the $+0.67$ peak gap lies inside the development spread of $15.6$--$25.7$ best-so-far reward observed across development repeats of adaptive-$\delta$ + PUCT on the C910 unit.

\item {\bf Theory assumptions and proofs}\\
Question: For each theoretical result, does the paper provide the full set of assumptions and a complete (and correct) proof?\\
Answer: \answerNA{}\\
Justification: The paper does not prove new theorems. \cref{sec:method} reuses the entropic policy-gradient objective and binary-search $\betastar$ from TTT-Discover \citep{yuksekgonul2026discover} and adds an empirical control rule (\cref{sec:adaptive-delta-variant}); we do not claim convergence or optimality results.

\item {\bf Experimental result reproducibility}\\
Question: Does the paper fully disclose all the information needed to reproduce the main experimental results of the paper to the extent that it affects the main claims and/or conclusions of the paper?\\
Answer: \answerYes{}\\
Justification: \cref{sec:experiments} and \cref{app:hyperparams} specify the base model (RTLLM~v2.0: Qwen3-8B + \texttt{ttt-rtl-sft} step 18; C910 LZA: raw Qwen3-8B with no SFT), training framework (\texttt{verl}), $100$-step budget, $B=4$ parents, $n\in\{4,8\}$ rollouts, PUCT $c=1.0$, pool cap $C_{\max}=500$, reward weights (\cref{eq:reward}), KL-budget controller hyperparameters (\cref{tab:adaptive_hparams}), PDKs (Nangate~45\,nm, Sky130 HD), tools (Yosys, OpenSTA, iverilog), seed ($42$), and prompt templates (\cref{app:prompts}). The artifact (to be released on GitHub) ships per-rollout logs, all generated Verilog, synthesis scripts, liberty-file manifests, and machine-readable result tables; re-evaluating any reported ratio from the released Verilog requires no retraining.

\item {\bf Open access to data and code}\\
Question: Does the paper provide open access to the data, code, and instructions needed to faithfully reproduce the main experimental results, as described in supplemental material?\\
Answer: \answerYes{}\\
Justification: All datasets and PDKs are public: RTLLM~v2.0 \citep{lu2024rtllm,liu2025openllmrtl}, the open-source XuanTie C910 RTL \citep{chen2020xuantie}, Nangate~45\,nm and Sky130. The artifact (to be released on GitHub under Apache 2.0) contains synthesis scripts, generated Verilog, prompts, and result CSVs; the training code and SFT checkpoint will be released alongside.

\item {\bf Experimental setting/details}\\
Question: Does the paper specify all the training and test details (e.g., data splits, hyperparameters, how they were chosen, type of optimizer, etc.) necessary to understand the results?\\
Answer: \answerYes{}\\
Justification: Per-run hyperparameters are in \cref{sec:experiments} (training config, sampling budget, baselines, synthesis flow, PDK choice) and \cref{app:hyperparams,tab:adaptive_hparams}. We also disclose the hyperparameter selection protocol: TTT-Discover defaults are inherited unmodified except for the adaptive KL-budget controller, whose thresholds were fixed once on the C910 LZA pilot and held constant across all $49$ RTLLM~v2.0 designs and the second C910 unit (\cref{sec:adaptive-delta-variant}, \cref{app:kl_budget_multiseed}).

\item {\bf Experiment statistical significance}\\
Question: Does the paper report error bars suitably and correctly defined or other appropriate information about the statistical significance of the experiments?\\
Answer: \answerYes{}\\
Justification: The main RTLLM~v2.0 and C910 main-table numbers are single-seed, matching all three published agent baselines (none of EvolVE / VeriAgent / REvolution report seed variance), and we say so explicitly in \cref{sec:experiments} and \cref{tab:ablations} (``All rows are single-seed''). For the adaptive vs.\ fixed $\delta$ contrast, \cref{app:kl_budget_multiseed} reports a four-seed paired replication on LZA \texttt{simd\_half} with mean $\pm$ std and paired $t$-test $p$-values for four metrics (\cref{tab:multiseed_lza}), an exhaustive enumeration over all $24$ adaptive-to-fixed pairings, leave-one-out checks, and a single-seed second-task replication on \texttt{fadd\_close\_s0\_d}. The negative-significance results are reported alongside the positive variance-reduction result.

\item {\bf Experiments compute resources}\\
Question: For each experiment, does the paper provide sufficient information on the computer resources (type of compute workers, memory, time of execution) needed to reproduce the experiments?\\
Answer: \answerYes{}\\
Justification: All experiments use A800 GPUs.  RTLLM~v2.0: each of the $49$ designs runs on $2$ nodes $\times$ $4$ A800 GPUs ($8$ GPUs total), wall-clock ${\sim}2$\,h per design (${\sim}16$ GPU-h per design); the full $49$-design sweep is ${\sim}780$ GPU-h.  C910 LZA ablations: each row runs on $1$ node $\times$ $8$ A800 GPUs, wall-clock ${\sim}4.5$\,h per row (${\sim}36$ GPU-h per row); the $8$ ablation rows in \cref{tab:ablations} total ${\sim}320$ GPU-h.  The seed and replication protocol is in \cref{sec:experiments} (``Compute and seed protocol'').

\item {\bf Code of ethics}\\
Question: Does the research conducted in the paper conform, in every respect, with the NeurIPS Code of Ethics?\\
Answer: \answerYes{}\\
Justification: The work uses publicly released model weights (Qwen3-8B), public benchmarks (RTLLM~v2.0), and the open-source XuanTie C910 RTL under their respective permissive licenses. No human subjects, no crowdsourcing, no scraped or private data are involved. The synthesized RTL is hardware-design output, not personal data.

\item {\bf Broader impacts}\\
Question: Does the paper discuss both potential positive societal impacts and negative societal impacts of the work performed?\\
Answer: \answerYes{}\\
Justification: TTT-RTL targets PPA reduction in fixed-function hardware modules, which can reduce energy and silicon area for deployed designs. Negative impacts are limited: the framework requires both an LLM and a full EDA flow, so it does not lower the barrier to malicious hardware generation beyond what is already achievable with standard EDA tools. We do not foresee dual-use risk that would warrant a release-time gating mechanism beyond the standard model card.

\item {\bf Safeguards}\\
Question: Does the paper describe safeguards that have been put in place for responsible release of data or models with a high risk for misuse (e.g., pretrained language models, image generators, or scraped datasets)?\\
Answer: \answerNA{}\\
Justification: We release fine-tuned Verilog-domain weights of an existing public base model (Qwen3-8B) plus per-design rollout logs and synthesis scripts. The artifact does not contain scraped data, pretrained generative models for natural-language or image content, or any content with elevated misuse risk relative to the base model.

\item {\bf Licenses for existing assets}\\
Question: Are the creators or original owners of assets (e.g., code, data, models), used in the paper, properly credited and are the license and terms of use explicitly mentioned and properly respected?\\
Answer: \answerYes{}\\
Justification: All third-party assets are cited in-text and respected in license: Qwen3-8B \citep{yang2025qwen3} (Apache 2.0), \texttt{verl} \citep{sheng2024hybridflow} (Apache 2.0), Yosys \citep{wolf2013yosys} (ISC), OpenSTA (GPLv3), iverilog (GPLv2), Nangate Open Cell Library 45\,nm (Apache-style), Sky130 PDK (Apache 2.0), RTLLM~v2.0 \citep{lu2024rtllm,liu2025openllmrtl} (MIT), the XuanTie C910 RTL \citep{chen2020xuantie} (Apache 2.0), and baseline numbers from \citet{ping2026coevo} (released under MIT). Per-asset versions are listed in \cref{sec:experiments,app:hyperparams}.

\item {\bf New assets}\\
Question: Are new assets introduced in the paper well documented and is the documentation provided alongside the assets?\\
Answer: \answerYes{}\\
Justification: New assets are: (a) the TTT-RTL training code, (b) the \texttt{ttt-rtl-sft} SFT checkpoint, (c) generated Verilog for all $49 \times 4 = 196$ method$\times$design rollout pools, and (d) the \texttt{reference\_ppa\_measured.csv} flow-sanity table (\cref{app:flow_sanity}). The artifact contains a top-level README with the directory layout, run instructions, license (Apache 2.0), and per-design provenance metadata. \cref{app:hyperparams} doubles as a model card for the SFT checkpoint.

\item {\bf Crowdsourcing and research with human subjects}\\
Question: For crowdsourcing experiments and research with human subjects, does the paper include the full text of instructions given to participants and screenshots, if applicable, as well as details about compensation (if any)?\\
Answer: \answerNA{}\\
Justification: No crowdsourcing or human-subjects studies were conducted. All evaluation signals come from automated EDA tools (Yosys, OpenSTA, iverilog) on public Verilog inputs.

\item {\bf Institutional review board (IRB) approvals or equivalent for research with human subjects}\\
Question: Does the paper describe potential risks incurred by study participants, whether such risks were disclosed to the subjects, and whether Institutional Review Board (IRB) approvals (or an equivalent approval/review based on the requirements of your country or institution) were obtained?\\
Answer: \answerNA{}\\
Justification: No human-subjects research was performed; IRB review does not apply.

\item {\bf Declaration of LLM usage}\\
Question: Does the paper describe the usage of LLMs if it is an important, original, or non-standard component of the core method development in this research?\\
Answer: \answerYes{}\\
Justification: An LLM is the central object of study---the policy that TTT-RTL fine-tunes per design. The base model (Qwen3-8B), the SFT corpus (\texttt{ttt-rtl-sft}, step 18), the training framework (\texttt{verl}), the prompt templates (\cref{app:prompts}), and all training-time hyperparameters (\cref{tab:hyperparams,tab:adaptive_hparams}) are documented in \cref{sec:experiments} and the appendix. LLMs were not used as a tool for paper writing or experiment design beyond standard authorial assistance, and any such use does not affect the reported methodology, results, or originality.

\end{enumerate}

\end{document}